\documentclass[]{template}

\usepackage{latexsym}
\usepackage{nicefrac}
\usepackage{multirow}
\usepackage{tikz}
\usepackage{xspace}
\usepackage{float}
\usepackage{subcaption}
\usepackage{wrapfig}
\usepackage{bm}
\usepackage{mathtools}
\usepackage{ragged2e}
\usepackage{adjustbox}

\newcolumntype{L}{>{\RaggedRight\hangafter=1\hangindent=0em}X}

\setboolean{logo}{true}

\hypersetup{
  colorlinks=true,
  linkcolor=compassblue,
  citecolor=compassblue,
  urlcolor=compassblue,
}

\usepackage[capitalize,noabbrev]{cleveref}
\crefname{section}{§}{§§}
\Crefname{section}{§}{§§}

\newcommand{\benchname}{\textsc{RNG-Bench}\xspace}

\newcommand{\cmark}{\textcolor{green!55!black}{\ding{51}}}
\newcommand{\xmark}{\textcolor{red!70!black}{\ding{55}}}
\newcommand{\pmark}{\textcolor{orange!80!black}{$\boldsymbol{\sim}$}}


\definecolor{matchpurple}{HTML}{8B7CF6}
\definecolor{matchamber}{HTML}{F59E0B}
\definecolor{matchgray}{HTML}{B8B8B8}
\newcommand{\modelname}[2]{#1\hspace{0.45em}#2}
\newcommand{\matchbar}[3]{%
  \begin{tikzpicture}[baseline=0.45ex, x=1cm, y=1cm]
    \def\barmax{1.05}
    \pgfmathsetmacro{\barwidth}{min(max(#2/#3,0),1)*\barmax}
    \fill[#1!14, rounded corners=1pt] (0,0) rectangle (\barmax,0.28);
    \shade[left color=#1!60, right color=#1!18, rounded corners=1pt]
      (0,0) rectangle ({\barwidth},0.28);
    \node at ({\barmax/2},0.14) {\fontsize{6.5}{7.5}\selectfont #2};
  \end{tikzpicture}%
}
\newcommand{\matchcellbar}[3]{%
  \begin{tikzpicture}[baseline=0.45ex, x=1cm, y=1cm]
    \def\barmax{1.05}
    \pgfmathsetmacro{\barwidth}{min(max(#2/#3,0),1)*\barmax}
    \fill[#1!14, rounded corners=1pt] (0,0) rectangle (\barmax,0.28);
    \shade[left color=#1!62, right color=#1!24, rounded corners=1pt]
      (0,0) rectangle ({\barwidth},0.28);
    \node at ({\barmax/2},0.14) {\fontsize{6.5}{7.5}\selectfont #2};
  \end{tikzpicture}%
}

\newcommand{\modelicon}[1]{%
  \makebox[0.42cm][c]{\raisebox{-0.22\height}{\includegraphics[width=0.30cm]{#1}}}%
}

\newcommand{\openaiicon}{\modelicon{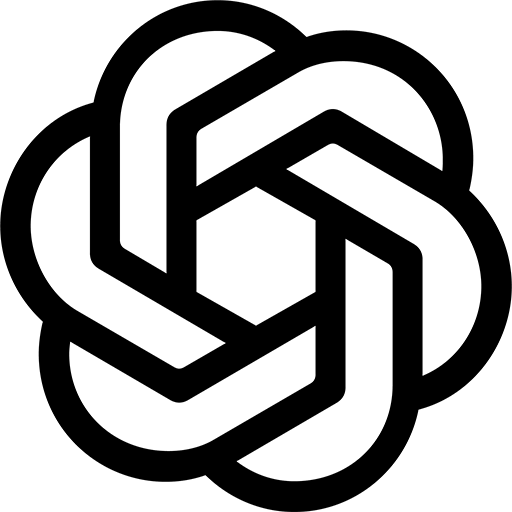}}
\newcommand{\geminiicon}{\modelicon{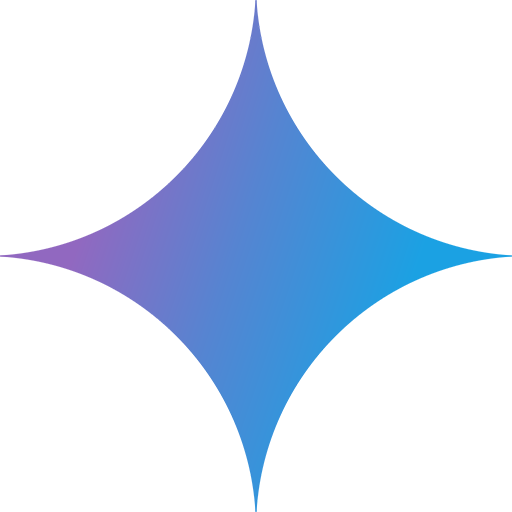}}
\newcommand{\qwenicon}{\modelicon{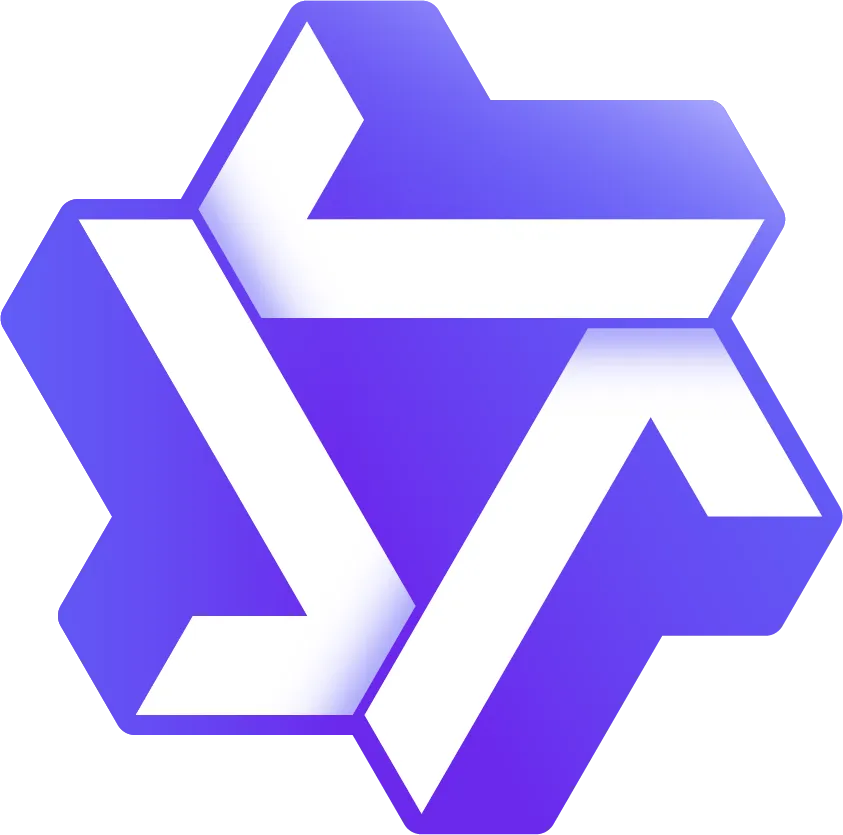}}
\newcommand{\seedicon}{\modelicon{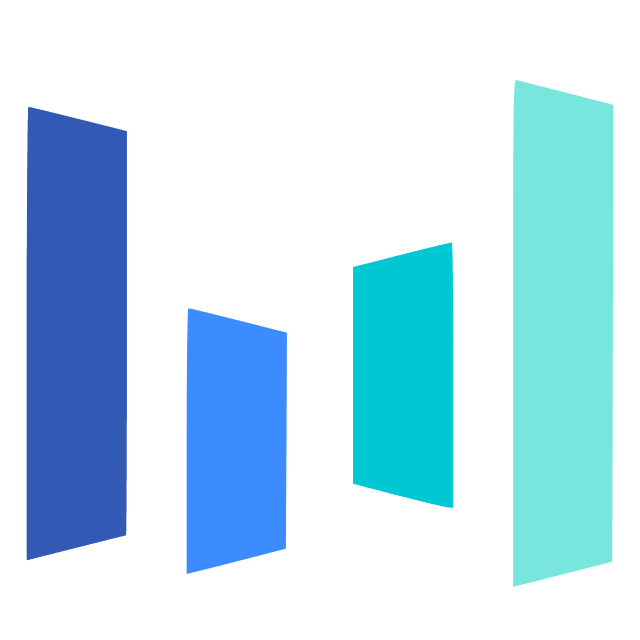}}
\newcommand{\kimiicon}{\modelicon{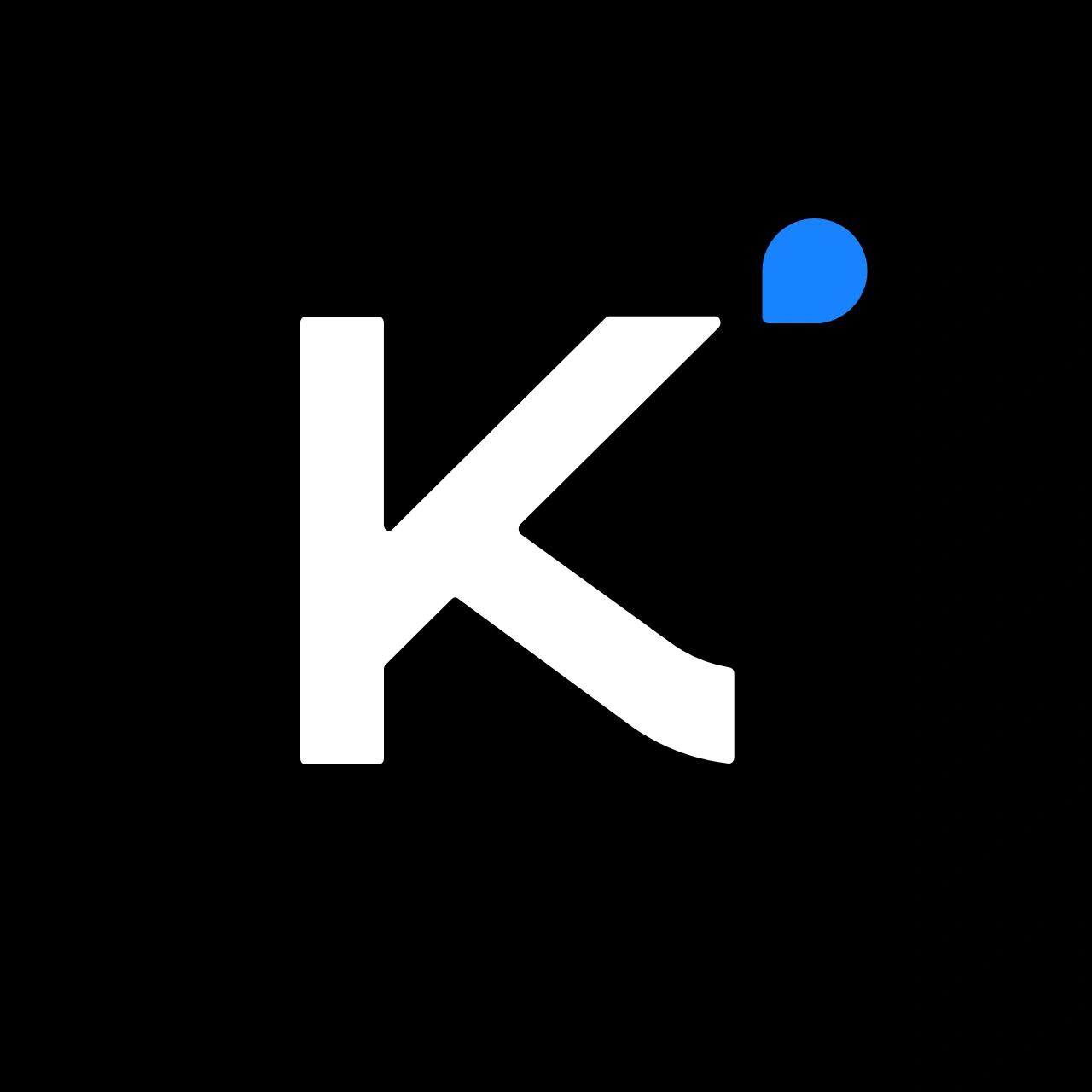}}

\renewcommand{\paragraph}[1]{\vspace{1mm}\noindent\textbf{#1}}

\newcommand\blfootnote[1]{%
  \begingroup
  \renewcommand\thefootnote{}\footnote{#1}%
  \addtocounter{footnote}{-1}%
  \endgroup
}

\title{Beyond the Current Observation: Evaluating Multimodal Large Language Models in Controllable Non-Markov Games}

\author[1,2,3]{Shengyuan Ding\textsuperscript{*}}
\author[1]{Xilin Wei\textsuperscript{*}}
\author[4]{Xinyu Fang\textsuperscript{*}}
\author[5]{Haodong Duan\textsuperscript{\textdagger}}
\author[3,5]{\protect\\ Dahua Lin}
\author[2]{Jiaqi Wang\textsuperscript{\textdagger}}
\author[3]{Yuhang Zang\textsuperscript{\textdagger}}

\affil[1]{Fudan University}
\affil[2]{Shanghai Innovation Institute}
\affil[3]{Shanghai Artificial Intelligence Laboratory}
\makeatletter\renewcommand\AB@affilsep{,\protect\\\protect\Affilfont}\makeatother
\affil[4]{Zhejiang University}
\affil[5]{The Chinese University of Hong Kong}

\begin{abstract}

Deploying multimodal foundation models as closed-loop policies increasingly requires conditioning actions on observations that are no longer visible. 
However, existing benchmarks either expose the full state, conflate hidden-state reconstruction with other agent skills, or test recall only after an episode has ended. 
We introduce \textbf{\benchname{}}~(\textbf{Reconstructive Non-Markov Games}), 
a benchmark suite designed to isolate a base model’s ability to reconstruct past observations and act on them during multi-step interaction. 
\benchname{} includes two complementary games: \textbf{Matching Pairs}, where card identities briefly revealed at specific locations must later be recalled, 
and \textbf{3D Maze}, where egocentric views must be integrated into a spatial map. 
Both games are evaluated under a unified harness with three controlled difficulty axes: grid size, visual pattern, and observation modality. 
The benchmark further introduces a head-to-head \textbf{duel} protocol to control for instance-level variance and a \textbf{Memory Gap} metric that disentangles forgetting from poor action selection. 
The hardest configurations require contexts of roughly 128K tokens and 350 image inputs per episode, and remain far from saturated by frontier MLLMs. 
Memory Gap analysis shows that most residual errors stem from forgetting earlier observations rather than from suboptimal decision making. 
Finally, fine-tuning Qwen3.5-9B on optimal-policy rollouts and filtered model demonstrations improves performance on \benchname{} and transfers to existing benchmarks without degrading general multimodal capability.

\end{abstract}

\begin{document}

\maketitle
\blfootnote{\textsuperscript{*}Equal contribution. \textsuperscript{\textdagger}Corresponding authors.}

\section{Introduction}
\label{sec:introduction}

\begin{figure}[tbp]
  \centering
  \includegraphics[width=\textwidth]{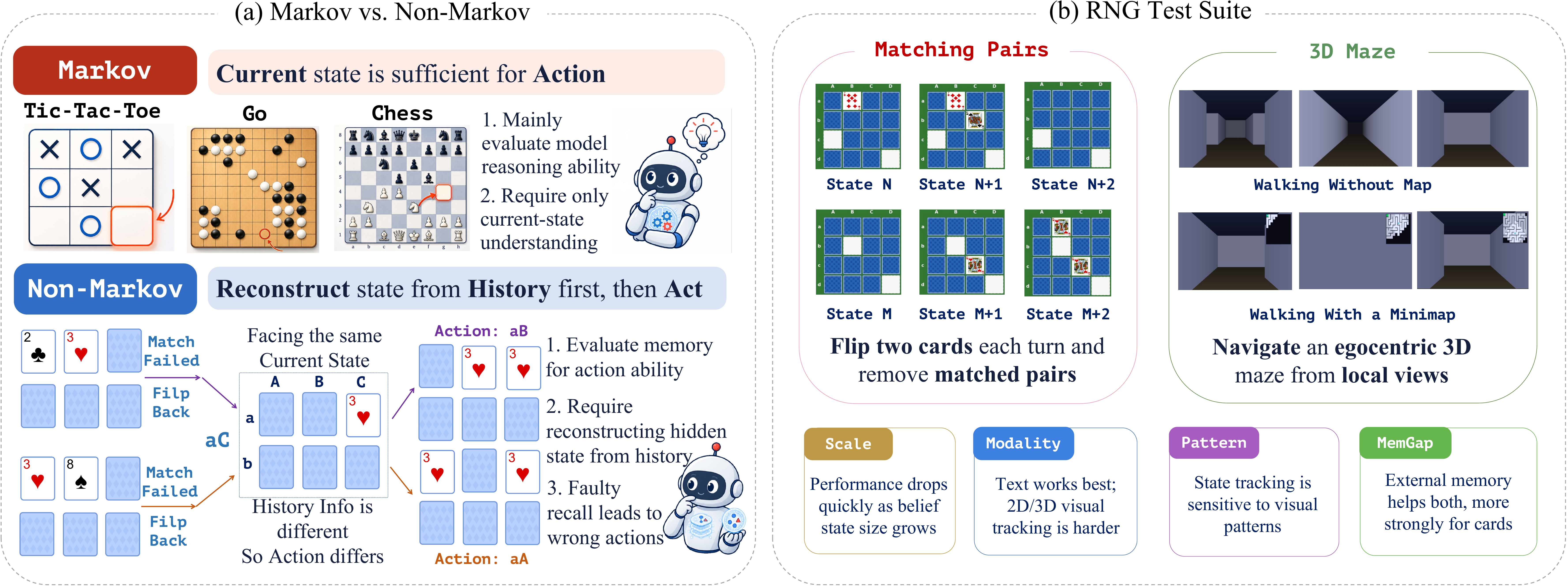}
  \caption{
  (a) Markov games are determined by the current state, while \benchname{} is non-Markov. (b) Two environments, Matching Pairs and 3D Maze, evaluated along three controlled axes (scale, visual pattern, and observation modality) with a Memory Gap diagnostic that isolates forgetting from action selection.
  }
  \label{fig:teaser}
\end{figure}

\begin{table}[tbp]
\centering
\small
\setlength{\tabcolsep}{4pt}
\renewcommand{\arraystretch}{1.05}
\adjustbox{max width=\textwidth}{%
\begin{tabular}{l l c c c c c c r r}
\toprule
\textbf{Benchmark} & \textbf{Domain} & \textbf{Eval} & \textbf{MM} & \textbf{Closed-loop} & \textbf{NM-Focus} & \textbf{Multi-pl.} & \textbf{Scal.-Diff.} & \textbf{Max Ctx} & \textbf{Max \#Img} \\
\midrule
\multicolumn{10}{l}{\emph{(i) Fully-visible game benchmarks}} \\
GameBench~\citep{costarelli2024gamebench}   & 9 strategic board / card games                  & Agent & \pmark & \cmark & \xmark & \cmark & \xmark & $6^\dagger$  & 1 \\
GTBench~\citep{duan2024gtbench}             & 10 game-theory games                            & Agent & \xmark & \cmark & \xmark & \cmark & \xmark & $8^\dagger$  & 0 \\
\midrule
\multicolumn{10}{l}{\emph{(ii) Agent / multi-environment suites with hidden information (bundled with other skills)}} \\
AgentBench~\citep{liu2023agentbench}        & Multi-task agent suite (8 envs)                 & Agent & \xmark & \cmark & \xmark & \xmark & \xmark & $12^\dagger$ & 0 \\
SmartPlay~\citep{wu2024smartplay}           & 6 text games (RPS, Bandit, Hanoi, \ldots)       & Agent & \xmark & \cmark & \xmark & \xmark & \cmark & $6^\dagger$  & 0 \\
AvalonBench~\citep{light2023avalonbench}    & Social deduction (hidden roles)                 & Agent & \xmark & \cmark & \xmark & \cmark & \xmark & $3.5^\dagger$ & 0 \\
BALROG~\citep{paglieri2024balrog}           & 6 RL game envs (incl.\ POMDPs: NetHack, \ldots) & Base  & \cmark & \cmark & \xmark & \xmark & \pmark & $16^\dagger$ & 1 \\
LMGame-Bench~\citep{hu2025lmgamebench}      & 6 video games + scaffolds                       & Both  & \cmark & \cmark & \xmark & \xmark & \xmark & 20           & 1 \\
GameWorld~\citep{ouyang2026gameworld}       & 34 browser games, 170 tasks                     & Agent & \cmark & \cmark & \xmark & \xmark & \pmark & $8^\dagger$  & 3 \\
MACHIAVELLI~\citep{pan2023machiavelli}      & 134 text-adventure games                        & Base  & \xmark & \cmark & \xmark & \xmark & \xmark & $2^\dagger$  & 0 \\
clembench~\citep{chalamalasetti2023clembench} & Dialogue games                                & Base  & \xmark & \cmark & \xmark & \cmark & \cmark & $6^\dagger$  & 0 \\
TextArena~\citep{guertler2025textarena}     & 100+ text games (TrueSkill rating)              & Base  & \xmark & \cmark & \xmark & \cmark & \pmark & $32^\dagger$ & 0 \\
\midrule
\multicolumn{10}{l}{\emph{(iii) Long-context \& memory benchmarks (game-based)}} \\
EMemBench~\citep{li2026emembench}           & Text + visual games (episodic-memory QA)        & Both  & \cmark & \xmark & \xmark & \xmark & \xmark & $20^\dagger$ & 4 \\
\midrule
\rowcolor{gray!12}
\textbf{\benchname{} (ours)} & \textbf{Non-Markov games (2D card + 3D maze)} & \textbf{Base} & \cmark & \cmark & \cmark & \cmark & \cmark & \textbf{128} & \textbf{350} \\
\bottomrule
\end{tabular}%
}
\caption{\benchname{} vs.\ prior benchmarks, grouped as in §\ref{sec:introduction}: (i) fully-visible games, (ii) agent suites that mix hidden information with other skills, (iii) game-based memory benchmarks. \textbf{Eval}: raw model (\emph{Base}) vs.\ wrapped harness (\emph{Agent}). \textbf{MM}: multimodal observations. \textbf{Closed-loop}: per-step action vs.\ post-hoc QA. \textbf{NM-Focus}: non-Markov recall as the central axis. \textbf{Multi-pl.}: duel or multi-agent protocol. \textbf{Scal.-Diff.}: controllable difficulty knobs. \textbf{Max Ctx} and \textbf{Max \#Img}: max per-prompt tokens and images. $\sim$ = partial; $^\dagger$ = estimated from code.}
\label{tab:bench_compare}
\end{table}

In long-horizon interaction, the correct action often depends on observations from several turns earlier rather than on the current view. 
A single recall error can change subsequent observations and compound throughout the episode.
We refer to this regime as \emph{Non-Markov}: 
the current observation alone is insufficient for optimal action, 
so the model must infer the relevant hidden state from its history before acting.
As multimodal models are deployed in closed-loop settings such as embodied control and multi-turn tool use, 
this ability is becoming critical alongside reasoning over visible inputs.

Fig.~\ref{fig:teaser}(a) makes the contrast concrete: in Markov games such as Go or chess the visible board determines the next move, whereas in Matching Pairs two boards with the same visible state can require different actions when their histories differ, so the visible state is not a sufficient statistic. This places two demands on the model: it must retain identities and locations across many turns and re-bind them to fading visual evidence, and it must tolerate that any recall error is causal, altering the next observation rather than only the final score.

Existing benchmarks fall into three families, none of which isolates this regime (Tab.~\ref{tab:bench_compare}). Fully observed games such as Go and chess~\citep{silver2018general} and strategic-reasoning suites~\citep{costarelli2024gamebench, duan2024gtbench} reward search and planning over a state that is already visible and do not require recall of earlier observations. Agent and multi-environment suites~\citep{liu2023agentbench, wu2024smartplay, paglieri2024balrog, light2023avalonbench, hu2025lmgamebench, ouyang2026gameworld, pan2023machiavelli, chalamalasetti2023clembench, guertler2025textarena} do include hidden information, but bundle it with exploration, rule discovery, and free-form action, so the four confounds named above remain entangled with memory at the level of episode outcomes. Long-context and memory benchmarks~\citep{bai2023longbench, hsieh2024ruler, maharana2024locomo, wu2024longmemeval, li2026emembench, creation_2025_ICCV} do isolate recall, but probe it post-hoc: the model reads a trajectory and answers a question once, a \emph{remember-to-answer} setting in which a recall error has no effect on subsequent inputs. Our regime is instead \emph{remember-to-act}, where each recall feeds back into the next observation.
We introduce \benchname{} (Reconstructive Non-Markov Games), a benchmark designed to isolate \emph{remember-to-act} along controlled axes, instantiated as two complementary closed-loop games. \textbf{Matching Pairs} is a card game that each symbol is revealed for a single turn and must later be recalled by location, isolating \emph{static, categorical} hidden state. \textbf{3D Maze} is an egocentric navigation task that corridors leave the field of view as the player moves and must be reassembled into a map, isolating \emph{dynamic, spatial} hidden state. Both games run in a closed loop: the model issues one action per turn and a faulty recall becomes a wrong move that reshapes the remaining observations, in contrast to benchmarks that score a single post-hoc answer. We vary difficulty along controlled axes: grid or map size, visual pattern, and observation modality (text or image), while rule understanding is held fixed by in-prompt rules and action formatting by a strict parser, so a drop along any axis is attributable to the axis rather than to a parallel confound. To analyze residual failures, we compare each model against an oracle condition that injects the true hidden state at every step. The score gap between the two is our \emph{Memory Gap}, separating forgetting from decision making given the correct state.

\benchname{} leaves substantial headroom for current models. On image Matching Pairs at $10{\times}10$, GPT-5.4~\cite{gpt5.4} matches 62.3\% of pairs and Qwen3.5-397B~\cite{qwen3.5} reaches 25.3\%, and across 16 head-to-head duels Gemini-3.1-Pro~\citep{gemini3.1pro} wins every matchup. For reference, an optimal policy uses roughly 60\% fewer moves per matched pair than the strongest model (3.24 vs.\ 8.01).
On 3D Maze at $13{\times}13$ (mean optimal path 60 steps), Gemini-3.1-Pro obtains the best result with 50.0\% SR and 49.7\% GS, while GPT-5.4 and Seed-2.0-Lite each reach 20.0\% SR, Kimi-K2.5 reaches 10.0\% SR, and Qwen3.5-397B reaches 0.0\% SR, a ranking that diverges from Matching Pairs and points to a distinct hidden-state demand.
The Memory Gap is consistent with forgetting accounting for a large share of the residual error rather than decision making given the correct state, and we further show that supervised fine-tuning on simulator rollouts from \benchname{} narrows this gap and transfers to external memory and spatial benchmarks (Sec.~\ref{sec:training_strategy}).


\noindent \textbf{Contributions:} \textbf{(1)} We release \benchname{}, two games (Matching Pairs and 3D Maze) under a unified closed-loop harness with a duel protocol. \textbf{(2)} We evaluate leading multimodal models along three controlled axes (grid or map size, visual pattern, and observation modality) and introduce a Memory Gap metric that attributes failures to forgetting earlier observations. \textbf{(3)} We construct training data from optimal-policy rollouts and filtered model demonstrations; fine-tuning Qwen3.5-9B improves \benchname{} performance and transfers to existing benchmarks~\citep{li2026emembench,ren2025vgrp} without regressing on general capability.
\section{Related Work}
\label{sec:related}


\noindent\textbf{Game Benchmarks.} Game benchmarks evaluate reasoning, planning, and multimodal action in interactive settings. AgentBench covers diverse agent tasks \citep{liu2023agentbench}, while GameBench, BALROG, and GameWorld use games to test strategic reasoning and multimodal game play \citep{costarelli2024gamebench,paglieri2024balrog,ouyang2026gameworld}. However, these broad benchmarks make memory-specific failures hard to isolate, as errors may come from perception, rule understanding, exploration, planning, or action formatting. In contrast, our environments make hidden state explicit and controllable, allowing Matching Pairs and 3D Maze to test whether models can recover information that is invisible and use it for later actions.

\noindent\textbf{Long-Context and Retrieval Benchmarks.} Long-context benchmarks study how well models use information distributed across extended inputs. LongBench, L-Eval, M4LE, and Ada-LEval provide broad long-context evaluation suites \citep{bai2023longbench,an2024leval,kwan2024m4le,wang2024adaleval}, while Lost in the Middle, RULER, NoLiMa, HELMET, and LongBench v2 probe position sensitivity, literal-match shortcuts, and realistic long-context reasoning \citep{liu2024lostmiddle,hsieh2024ruler,modarressi2025nolima,yen2025helmet,bai2025longbenchv2}. Retrieval and reading benchmarks such as MS MARCO, TriviaQA, KILT, and BEIR test whether models can locate relevant evidence from available sources \citep{nguyen2016msmarco,joshi2017triviaqa,kwiatkowski2019naturalquestions,yang2018hotpotqa,petroni2021kilt,thakur2021beir}, and MMNeedle extends this to long visual contexts \citep{wang2024mmneedle}. These show that placing evidence in a context window does not guarantee robust use of it. Our setting differs because long context is produced by interaction rather than given as a fixed input, and key evidence may be seen only once before it becomes useful later.
\section{Benchmark Design}
\label{sec:benchmark_design}

\subsection{Problem Formulation}
\label{sec:problem_formulation}

We model each benchmark instance as a \emph{Partially Observable Markov Decision Process (POMDP)} $(\mathcal{S}, \mathcal{O}, \mathcal{A}, T, Z, R)$~\citep{kaelbling1998planning}. Here, $\mathcal{S}$, $\mathcal{O}$, and $\mathcal{A}$ denote the state, observation, and action spaces. The transition function $T$ specifies how the state changes after an action, the observation function $Z$ specifies what the agent can observe from the current state, and $R$ gives the reward. At step $t$, the agent receives the current observation $o_t$ and the in-context episode history $h_t=(o_1,a_1,\ldots,o_{t-1},a_{t-1},o_t)$, then selects action $a_t$. We evaluate models as \emph{history-based policies} $\pi(a_t \mid h_t)$ that directly use the raw in-context history, with no external belief module by default.

The POMDP framework covers both fully observed games (when $Z$ preserves all task-relevant state) and partially observed ones where history matters. A game is \emph{Markov} when the current observation suffices for optimal play: $\pi^*(a_t\mid o_t)=\pi^*(a_t\mid h_t)$ , $\forall h_t$. Fully observed board games such as chess and Go, when encoded with all rule-relevant variables, are typical examples. A game is \emph{non-Markov} when the current observation alone  is not sufficient to determine the optimal action. Equivalently, two different histories can lead to the same current observation but require different actions. Let $\mathcal{A}^*(h_t)$ denote the set of optimal actions under history $h_t$. Then a non-Markov game satisfies:
\vspace{-5pt}
\begin{equation}
\label{eq:non_markov}
\begin{split}
\exists\, h_t, \tilde{h}_t \quad \text{s.t.} \quad
& Z(s_t) = Z(\tilde{s}_t), \\
& \mathcal{A}^*(h_t) \neq \mathcal{A}^*(\tilde{h}_t).
\end{split}
\end{equation}
\vspace{-1em}

To act well in such games, the agent must build a \emph{internal belief state} $b_t=f(h_t)$ from its interaction history. This state is an internal summary of the hidden, task-relevant information that is no longer directly visible. If the internal belief state is accurate, the agent can act as if the relevant state were observable. If it is inaccurate or incomplete, the agent may revisit known states, repeat ineffective actions, ignore useful past observations, and make locally plausible but globally wrong decisions.

Our benchmark is designed to test this ability: whether a model can maintain an accurate internal belief state within its in-context episode history and use that belief to choose the next action. We refer to this ability as \emph{in-context state tracking for action}.

\subsection{Why These Environments}
\label{sec:why_environments}

\begin{figure}[tbp]
\centering
\includegraphics[width=\textwidth]{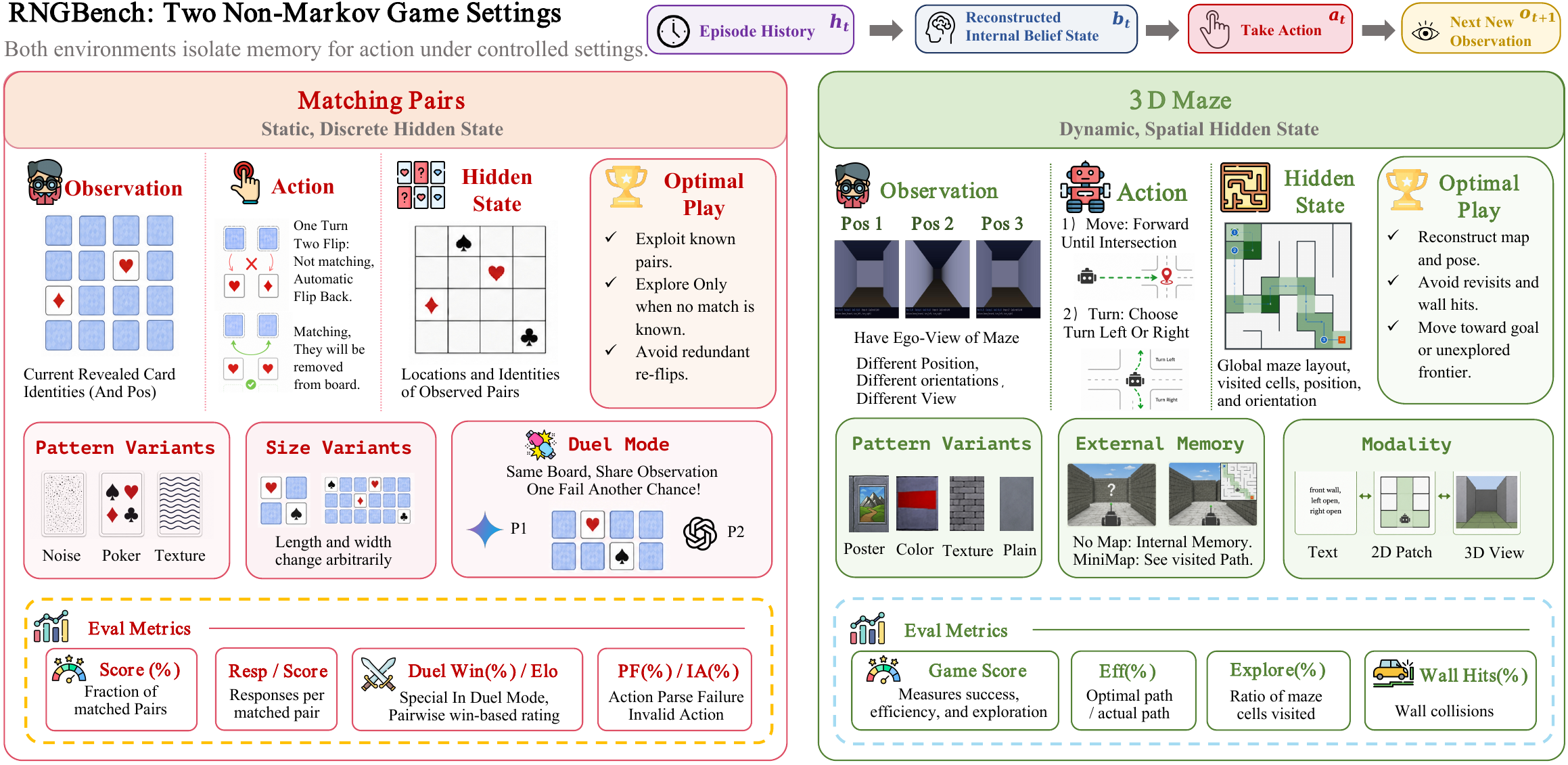}
\caption{\textbf{Two complementary environments for in-context state tracking.}
Matching Pairs tests static identity-location memory, while 3D Maze tests dynamic map construction from egocentric observations. Both use simple rules, scalable grids, and controllable visual settings to isolate belief-state tracking from other sources of difficulty.}
\label{fig:two_env}
\end{figure}

We choose Matching Pairs and 3D Maze because they are simple but diagnostic POMDP instances: the optimal action depends on information that appeared earlier in the episode but is no longer visible, so the model must reconstruct a belief state from its in-context history. The two games stress different hidden states (Fig.~\ref{fig:two_env}). Matching Pairs focuses on \emph{static, discrete, factual} state: card identities and positions are fixed, but each card is only briefly revealed, so the agent must remember which identity appeared at which location. 3D Maze focuses on \emph{dynamic, spatial, structural} state: position, orientation, visited cells, and local topology must be updated incrementally from egocentric views.

The two environments span three observation modalities (text, 2D image, and 3D rendering). The rules are simple, so failures reflect belief-state tracking rather than rule misunderstanding. Difficulty scales with grid size, increasing the hidden-state load without changing the task definition. Full trajectory logs expose fine-grained diagnostics (repeated flips, revisited cells, wall collisions), and targeted interventions (oracle state injection, scratchpads, minimap) can selectively remove the memory requirement to localize the bottleneck.
Both environments test a \emph{remember-to-act} ability: the model must use reconstructed belief state for immediate action, not merely recall information after the episode ends (\emph{remember-to-answer}).

\subsection{Environment Construction}
\label{sec:environment_construction}

\paragraph{Matching Pairs.}
A rectangular grid of size $R{\times}C$ is populated with $\frac{R \times C}{2}$ card pairs, each sharing a visual identity, all initially face-down. At each turn, the agent flips two cards. Matched cards are removed, while unmatched cards are turned back over. The game ends when all pairs are matched or a fixed response budget is exhausted. The hidden state is the set of previously revealed but currently hidden identity-location bindings. An optimal agent should use these bindings to find known pairs and avoid redundant re-flips.

We systematically vary \textbf{board size} (configurable to arbitrary $R{\times}C$), \textbf{observation modality} (image with controlled token count vs.\ text), \textbf{visual pattern} (e.g., ASCII glyphs, poker suits, textures, noise patterns, \ldots), \textbf{action feedback} (explicit flip result vs.\ no feedback), \textbf{CoT prompting} (allowed vs.\ direct action only), and \textbf{response budget} to isolate the contribution of each factor.



\paragraph{3D Maze.}
The agent navigates from the top-left start to the bottom-right goal in a procedurally generated grid maze. It default receives only an egocentric first-person rendering and the dialogue history, with no top-down map. The hidden state consists of the maze topology, visited cells, current position, and facing direction, all must be maintained incrementally from local views. The action space is \texttt{move\_forward}, \texttt{turn\_left} and \texttt{turn\_right}. 

Mazes are generated with a loop rate of $0.15$ (the fraction of extra openings added to the spanning tree), which introduces cycles that make simple wall-following less reliable. Each configuration is evaluated over five seeds with a step budget of $\max(80,\;4{\times}L^*)$, where $L^*$ is the shortest-path length. We vary \textbf{maze size} ($5{\times}5$ to $15{\times}15$), \textbf{minimap availability} (converting the task to approximately Markov), \textbf{ask-output prompting} (externalizing the spatial belief), and \textbf{history window} (3/5/10 turns vs.\ full history). These settings let us test how spatial belief tracking changes with scale, external memory, explicit belief reporting, and context length.

\subsection{Duel Protocol for Matching Pairs}
\label{sec:duel_protocol}



We introduce a duel protocol for Matching Pairs to compare models under the same hidden-state structure. Two models play on the same board with identical rules and take turns to flip cards. Each player observes the cards revealed by both itself and its opponent, but does not observe the opponent's reasoning. A successful match grants an extra turn, while a non-match passes the turn to the opponent. The player that removes more pairs wins.

Duel mode offers three advantages over single-agent evaluation. First, it controls for board randomness because both models face the same card layout. Second, it tests whether a model can use information revealed by the opponent's flips as part of its own belief state. Third, it gives a robustness ranking that complements single-agent scores. To control for first-mover bias, each pair of competitors plays the same board twice with swapped turn order. The aggregated duel result is used as a complementary robustness check for belief-state tracking under shared observations.

\subsection{Evaluation Metrics}
\label{sec:metrics}

Each environment has its own primary completion metric, supplemented by trajectory-level diagnostics that capture how failures occur. 
For Matching Pairs, we report \textbf{Score\%}, the fraction of pairs matched in each game, and \textbf{Resp./Score}, the average number of responses needed per matched pair, where lower is better. Parse failures and invalid actions are reported as additional diagnostics.
For 3D Maze, we report \textbf{Success Rate (SR)}, the fraction of episodes that reach the goal within the budget, and \textbf{Efficiency}, defined as $L^*/L_{\text{actual}}$ over successful episodes, where $L^*$ is the shortest-path length and $L_{\text{actual}}$ is the executed path length. We also report \textbf{Explore}, the ratio of visited cells, and \textbf{Walls}, the number of wall collisions. The primary scalar metric is \textbf{Game Score (GS)}:
\begin{equation}
\label{eq:game_score}
\text{GS} = \frac{\text{SR} + \text{SR}{\times}\text{Eff} + (1{-}\text{SR}){\times}\text{Explore}}{2},
\end{equation}
GS rewards successful completion, gives an efficiency bonus for successful episodes, and assigns partial credit for exploration when the agent fails to reach the goal. Its value lies in $[0,1]$.



\paragraph{Memory Gap.}
To separate belief-state tracking from action selection, we define an \emph{oracle} condition that provides the true hidden state $s_t$ into the prompt at each step. Let $S(m)$ and $S^*(m)$ denote the score of model $m$ under the normal and oracle conditions, respectively. Depending on the setting, $S$ can be SR, Score\%, or Efficiency. The Memory Gap is defined as
\begin{equation}
\label{eq:memorygap}
\text{MemoryGap}(m) = \left(1 - \frac{S(m)}{S^*(m)}\right) \times 100\,\%.
\end{equation}
A large Memory Gap points to internal belief state reconstruction as the main bottleneck, while a small gap suggests errors in action selection, rule understanding, or perception.

\section{Experiments}
\label{sec:experiments}
\begingroup
\renewcommand{\arraystretch}{1.35}
\setlength\extrarowheight{1.4pt}
\renewcommand{\matchcellbar}[3]{%
  \begin{tikzpicture}[baseline=0.45ex, x=1cm, y=1cm]
    \def\barmax{1.30}
    \pgfmathsetmacro{\barwidth}{min(max(#2/#3,0),1)*\barmax}
    \fill[#1!14, rounded corners=1pt] (0,0) rectangle (\barmax,0.38);
    \shade[left color=#1!62, right color=#1!24, rounded corners=1pt]
      (0,0) rectangle ({\barwidth},0.38);
    \node at ({\barmax/2},0.19) {\fontsize{8.5}{9.5}\selectfont #2};
  \end{tikzpicture}%
}
\begin{table}[tbp]
\setlength\tabcolsep{2.4pt}
\centering
\small
\adjustbox{max width=\textwidth}{%
\begin{tabular}{>{\raggedright\arraybackslash}m{3.05cm}|>{\centering\arraybackslash}m{0.90cm}>{\centering\arraybackslash}m{0.90cm}>{\centering\arraybackslash}m{1.75cm}>{\centering\arraybackslash}m{1.45cm}|>{\centering\arraybackslash}m{0.95cm}>{\centering\arraybackslash}m{1.60cm}>{\centering\arraybackslash}m{1.45cm}>{\centering\arraybackslash}m{1.45cm}>{\centering\arraybackslash}m{1.45cm}}
\toprule
\multirow{2}{*}{\textbf{Model Name}}
& \multicolumn{4}{c|}{\textbf{Matching Pairs $10{\times}10$}}
& \multicolumn{5}{c}{\textbf{3D Maze $13{\times}13$}} \\
& \textbf{PF\%}\,$\downarrow$ & \textbf{IA\%}\,$\downarrow$ & \textbf{Resp./Score}\,$\downarrow$ & \textbf{Score\%}\,$\uparrow$
& \textbf{SR\%}\,$\uparrow$ & \textbf{Explore\%}\,$\uparrow$ & \textbf{Walls}\,$\downarrow$ & \textbf{Eff.\%}\,$\uparrow$ & \textbf{GS\%}\,$\uparrow$ \\
\midrule
\modelname{\openaiicon}{GPT-5.4}
& \textbf{0.0} & 4.3 & \textbf{\matchcellbar{matchamber}{8.01}{45.0}} & \textbf{\matchcellbar{matchpurple}{62.3}{100.0}}
& 20.0 & \matchcellbar{matchgray}{32.3}{100.0} & \matchcellbar{matchgray}{3.2}{35.0} & \textbf{\matchcellbar{matchpurple}{75.7}{100.0}} & \matchcellbar{matchpurple}{30.5}{100.0} \\
\modelname{\geminiicon}{Gemini-3.1-Pro}
& 0.4 & 2.5 & \matchcellbar{matchamber}{10.00}{45.0} & \matchcellbar{matchpurple}{50.0}{100.0}
& \textbf{50.0} & \matchcellbar{matchgray}{36.4}{100.0} & \textbf{\matchcellbar{matchgray}{0.1}{35.0}} & \matchcellbar{matchpurple}{62.5}{100.0} & \textbf{\matchcellbar{matchpurple}{49.7}{100.0}} \\
\modelname{\seedicon}{Seed-2.0-Lite}
& 1.2 & 4.3 & \matchcellbar{matchamber}{11.57}{45.0} & \matchcellbar{matchpurple}{43.2}{100.0}
& 20.0 & \matchcellbar{matchgray}{19.4}{100.0} & \matchcellbar{matchgray}{16.6}{35.0} & \matchcellbar{matchpurple}{38.9}{100.0} & \matchcellbar{matchpurple}{21.7}{100.0} \\
\modelname{\kimiicon}{Kimi-K2.5}
& 1.8 & 2.8 & \matchcellbar{matchamber}{13.16}{45.0} & \matchcellbar{matchpurple}{38.0}{100.0}
& 10.0 & \matchcellbar{matchgray}{17.9}{100.0} & \matchcellbar{matchgray}{7.1}{35.0} & \matchcellbar{matchpurple}{61.1}{100.0} & \matchcellbar{matchpurple}{16.1}{100.0} \\
\modelname{\qwenicon}{Qwen3.5-397B}
& \textbf{0.0} & 3.0 & \matchcellbar{matchamber}{19.74}{45.0} & \matchcellbar{matchpurple}{25.3}{100.0}
& 0.0 & \matchcellbar{matchgray}{21.0}{100.0} & \matchcellbar{matchgray}{9.9}{35.0} & \matchcellbar{matchpurple}{0.0}{100.0} & \matchcellbar{matchpurple}{10.5}{100.0} \\
\bottomrule
\end{tabular}
}
\caption{\textbf{Main results on the two environments (single-player setting)}. We focus on two metrics: \emph{Score}, the fraction of matched pairs on Matching Pairs, and \emph{GS}, the aggregate 3D Maze score combining success rate, efficiency, and exploration. The remaining columns support analysis: \emph{PF}/\emph{IA} means parse-failure and invalid-action rates;  \emph{Resp./Score} reports responses per matched pair; \emph{SR}, \emph{Explore}, \emph{Walls}, and \emph{Eff} refers to success rate, exploration coverage, wall collisions, and path efficiency.}
\label{tab:main_results}
\end{table}
\endgroup

\renewcommand{\arraystretch}{1.08}
\begin{table}[tbp]
\setlength\tabcolsep{5pt}
\centering
\small
\adjustbox{max width=\columnwidth}{%
\begin{tabular}{l|cccc|c|c}
\toprule
\textbf{Model Name}
& \textbf{Win\%} & \textbf{W} & \textbf{T} & \textbf{L}
& \textbf{Score\%}
& \textbf{ELO} \\
\midrule
\modelname{\geminiicon}{Gemini-3.1-Pro} & 100.0 & 16 & 0 & 0 & \matchcellbar{matchpurple}{36.5}{100.0} & 1803 \\
\modelname{\openaiicon}{GPT-5.4}        & 50.0  & 7  & 2 & 7 & \matchcellbar{matchpurple}{25.3}{100.0} & 1492 \\
\modelname{\qwenicon}{Qwen3.5-397B}     & 46.7  & 7  & 1 & 8 & \matchcellbar{matchpurple}{18.0}{100.0} & 1476 \\
\modelname{\kimiicon}{Kimi-K2.5}        & 37.5  & 5  & 2 & 9 & \matchcellbar{matchpurple}{18.0}{100.0} & 1423 \\
\modelname{\seedicon}{Seed-2.0-Lite}    & 15.6  & 2  & 1 & 13 & \matchcellbar{matchpurple}{12.3}{100.0} & 1306 \\
\bottomrule
\end{tabular}
}
\caption{\textbf{Main results on Matching Pairs (duel setting)}.
Each model plays 16 games against the other four, aggregating both player orders over two board seeds.
}
\label{tab:match_dual_image_poker_rank}
\end{table}

All models are evaluated under a unified harness built on top of VLMEvalKit~\citep{duan2024vlmevalkit}, ensuring identical prompts, parsing, and metric computation across models. We first report main results, followed by diagnostic analyses (hidden-state scale, external memory, observation modality, and action-feedback text) to pinpoint where belief-state tracking degrades. Additional analyses appear in the appendix.

\subsection{Main Results}
\label{sec:main_results}

\noindent\textbf{Single-player setting.} Tab.~\ref{tab:main_results} reports the main results on both environments. Matching Pairs uses a $10{\times}10$ board (50 pairs) with image observations and the \texttt{noise} card theme. GPT-5.4 leads at 62.3\% with the lowest response cost per matched pair (8.01). Gemini-3.1-Pro follows at 50.0\%, and Qwen3.5-397B reaches 25.3\%. Parse failures (PF\%) and invalid actions (IA\%) stay below 5\% across models, so the gaps are not explained by output-format compliance.

The ranking changes on 3D Maze. We evaluate on $13{\times}13$ mazes with no minimap and a mean optimal path length of 60.0 steps. Gemini-3.1-Pro obtains the highest success rate (50.0\%) and game score (49.7\%), whereas GPT-5.4 reaches 20.0\% SR and 30.5\% GS despite leading on Matching Pairs. Seed-2.0-Lite matches GPT-5.4's SR but trails in GS, while Kimi-K2.5 and Qwen3.5-397B remain lower. This suggests that the two tasks stress different forms of hidden-state tracking: identity retention and pairwise retrieval in Matching Pairs, versus spatial belief updating and route planning in 3D Maze.


\noindent\textbf{Duel setting.} The duel mode is a complementary evaluation for Matching Pairs: two models alternate turns on the same board, and each player's flips expose card identities the opponent can exploit. Unlike the single-player setting, this tests whether a model can track both its own flips and the cards revealed by another model.

Tab.~\ref{tab:match_dual_image_poker_rank} reports head-to-head results across five models. Gemini-3.1-Pro wins all 16 games and tops the Elo ranking. GPT-5.4 and Qwen3.5-397B post similar win rates (50\% and 47\%), but GPT-5.4 averages more matched pairs per game (10.1 vs.\ 7.2) and a slightly higher Elo. Kimi-K2.5~\citep{team2026kimi} wins 38\% of games, while Seed-2.0-Lite~\citep{seed2_lite} wins 16\%.
%
The duel ranking partially differs from the single-player order. GPT-5.4 drops from first to second and Gemini-3.1-Pro rises from second to first. Because Win rate (\%) in Tab.~\ref{tab:match_dual_image_poker_rank} aggregates over swapped player orders, the shift is unlikely to be a first- or second-mover bias. Instead, Gemini appears better at using card identities revealed by the opponent and converting them into consecutive matching turns. This strategic advantage helps explain its top duel performance and suggests stronger perception and retention of image-based information over long interaction histories.


\subsection{Diagnostic Analysis}
\label{sec:diagnostic_analysis}



\begin{figure}[tbp]
\centering
\begin{minipage}[t]{0.49\textwidth}
\centering
\includegraphics[width=\linewidth]{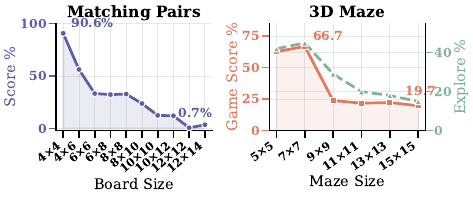}
\caption{\textbf{Hidden-state scale sweep.} \textit{Left}: Matching Pairs Score\% from $4{\times}4$ to $12{\times}14$. \textit{Right}: 3D Maze Game Score and Explore\% from $5{\times}5$ to $15{\times}15$. Performance \textbf{drops sharply} in both environments as the hidden state grows, pointing to belief-state maintenance rather than rule comprehension as the bottleneck.}
\label{fig:rq1_scale_sweep}
\end{minipage}\hfill
\begin{minipage}[t]{0.49\textwidth}
\centering
\includegraphics[width=\linewidth]{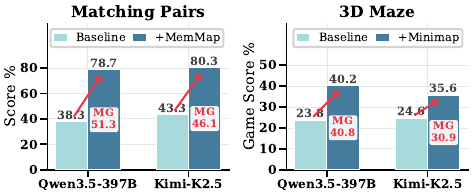}
\caption{\textbf{External-memory intervention.} Matching Pairs and 3D Maze with and without a memory map or minimap. Red arrows show MemGap in Eq.~\ref{eq:memorygap}. External memory doubles Matching Pairs but recovers a smaller share on 3D Maze, indicating spatial navigation couples hidden-state tracking with action planning.}
\label{fig:rq4_external_memory}
\end{minipage}
\end{figure}

\noindent\textbf{Performance drops sharply as the hidden state grows.} We test whether performance degrades as the hidden state grows while task rules stay fixed. In Matching Pairs, the hidden state scales with board size since the model must retain more card identities and locations; in 3D Maze, it scales with maze size and path complexity, requiring a larger spatial belief state over a longer action history. Fig.~\ref{fig:rq1_scale_sweep} shows a clear scale effect for Qwen3.5-397B in both environments: Matching Pairs Score\% drops from 90.6\% on $4{\times}4$ to 0.7\% on $12{\times}12$, while 3D Maze Game Score peaks at $7{\times}7$ and then declines from $9{\times}9$ onward, reaching 0.197 at $15{\times}15$. The parallel drop in Explore\% indicates that larger mazes degrade not only task success but also state-space coverage. The model can follow the rules at small scale, but its belief-state maintenance becomes unreliable as the latent state grows.

\renewcommand{\arraystretch}{1.15}
\begin{table}[tbp]
\centering
\setlength\tabcolsep{4.0pt}
\small
\adjustbox{max width=\linewidth}{%
\begin{tabular}{l|lc|lccc}
\toprule
\multirow{2}{*}{\textbf{Model Name}}
 & \multicolumn{2}{c|}{\textbf{Matching Pairs}}
 & \multicolumn{4}{c}{\textbf{3D Maze}} \\
 & \textbf{Modality} & \textbf{Score\%}\,$\uparrow$
 & \textbf{Modality} & \textbf{Eff.\%}\,$\uparrow$ & \textbf{Explore\%}\,$\uparrow$ & \textbf{GS\%}\,$\uparrow$ \\
\midrule
\multirow{3}{*}{\modelname{\qwenicon}{Qwen3.5-397B}}
 & Text         & \textbf{\matchcellbar{matchpurple}{100.0}{100.0}}
 & Text Symbolic   & 67.2 & \textbf{\matchcellbar{matchgray}{41.0}{100.0}} & \textbf{\matchcellbar{matchpurple}{70.9}{100.0}} \\
 & Image-ASCII  & \matchcellbar{matchpurple}{75.8}{100.0}
 & 2D Local Patch  & 27.5 & \matchcellbar{matchgray}{18.0}{100.0} & \matchcellbar{matchpurple}{20.1}{100.0} \\
 & Image-Noise  & \matchcellbar{matchpurple}{38.3}{100.0}
 & 3D Scene        & 21.9 & \matchcellbar{matchgray}{29.0}{100.0} & \matchcellbar{matchpurple}{23.7}{100.0} \\
\midrule
\multirow{3}{*}{\modelname{\kimiicon}{Kimi-K2.5}}
 & Text         & \textbf{\matchcellbar{matchpurple}{100.0}{100.0}}
 & Text Symbolic   & 74.9 & \textbf{\matchcellbar{matchgray}{38.0}{100.0}} & \textbf{\matchcellbar{matchpurple}{60.0}{100.0}} \\
 & Image-ASCII  & \matchcellbar{matchpurple}{72.5}{100.0}
 & 2D Local Patch  & 87.5 & \matchcellbar{matchgray}{17.0}{100.0} & \matchcellbar{matchpurple}{25.7}{100.0} \\
 & Image-Noise  & \matchcellbar{matchpurple}{43.3}{100.0}
 & 3D Scene        & 62.5 & \matchcellbar{matchgray}{24.0}{100.0} & \matchcellbar{matchpurple}{39.7}{100.0} \\
\bottomrule
\end{tabular}%
}
\caption{\textbf{Modality ablation.} Matching Pairs compares symbolic text, ASCII-style, and noise-pattern image cards; 3D Maze compares text-symbolic, 2D local patches, and 3D first-person views. Text-only modality dominate, indicating that visual recognition limits hidden-state tracking.}
\label{tab:rq2_modality}
\end{table}


\noindent\textbf{External memory recovers most of the gap on Matching Pairs but only part of it on 3D Maze.}
We test whether providing explicit external memory repairs hidden-state tracking. In Matching Pairs the intervention is a memory map of known cards; in 3D Maze it is a minimap with the visited-state summary. Fig.~\ref{fig:rq4_external_memory} reports both, annotated with our MemGap metric (Eq.~\ref{eq:memorygap}), which quantifies the residual deficit after external memory is supplied. On Matching Pairs, Qwen3.5-397B and Kimi-K2.5 roughly double their Score\% with the memory map, yielding MemGap values of 51.3 and 46.1. On 3D Maze the minimap closes a smaller share of the gap (MemGap 40.8 and 30.9). Kimi-K2.5's MemGap is uniformly lower than Qwen3.5-397B's, indicating that external memory helps Kimi-K2.5 less relative to its baseline. These results localize the Matching Pairs bottleneck primarily to hidden-state maintenance, while pointing to additional limitations beyond memory access in 3D Maze.


\noindent\textbf{Visual observations, not just memory length, drive the bottleneck.}
We disentangle visual perception from history-to-state tracking by varying the observation modality while holding the hidden state fixed. Matching Pairs compares symbolic text, ASCII-style image cards, and noise-pattern image cards; 3D Maze compares text-symbolic, 2D local patches, and 3D first-person views. Tab.~\ref{tab:rq2_modality} reports both environments side by side. Both Qwen3.5-397B and Kimi-K2.5 solve Matching Pairs perfectly under text but fall to 38.3\% and 43.3\% under noise-pattern images. The 3D Maze shows the same ordering, with text-symbolic Game Scores far above the 2D patch and 3D scene settings. That text-only configurations remain strong while image settings collapse indicates that visual recognition, not history length alone, limits hidden-state tracking in these models. Per-pair duel results on Matching Pairs are reported in Appendix.~\ref{sec:appendix_ablation}





\noindent\textbf{Removing the action-feedback text collapses Matching Pairs to near-chance.}\label{sec:diag_action_feedback}
We test whether the model's own action history is necessary when the rendered board already carries every visual change. Tab.~\ref{tab:ablation_action_feedback} strips the model's previous actions from the conversation history on Matching Pairs (image-noise), keeping only the sequence of board images; the model must re-infer which cells it flipped from the visual diff between consecutive boards. The effect is consistent across a leading closed-source and a leading open-source model: GPT-5.4's Score\% falls by roughly 75\% on both $8{\times}10$ and $10{\times}10$ (69.6$\to$15.0, 62.3$\to$15.3), and Qwen3.5-397B falls by about 70--75\% from a lower base. Parse failures and invalid actions remain at zero, so the collapse is not an output-format issue. Board images are in principle sufficient to recover each flip, yet neither model can close the loop from observation to belief update without an explicit textual record of its own actions. The action trace functions as a load-bearing channel for belief-state tracking rather than redundant decoration.

\begin{table}[tbp]
\centering

\begin{minipage}[t]{0.5\textwidth}
\centering
\renewcommand{\arraystretch}{1.12}
\setlength\tabcolsep{4pt}
\small
\adjustbox{max width=\linewidth}{%
\begin{tabular}{l|ccc|ccc}
\toprule
\multirow{2}{*}{\textbf{Board}}
& \multicolumn{3}{c|}{\modelname{\openaiicon}{GPT-5.4}}
& \multicolumn{3}{c}{\modelname{\qwenicon}{Qwen3.5-397B}} \\
& \textbf{w/\,Act.} & \textbf{w/o\,Act.} & $\boldsymbol{\Delta}_{\text{rel}}$\,(\%)
& \textbf{w/\,Act.} & \textbf{w/o\,Act.} & $\boldsymbol{\Delta}_{\text{rel}}$\,(\%) \\
\midrule
$8{\times}10$  & \matchcellbar{matchpurple}{69.6}{100.0}
               & \matchcellbar{matchgray}{15.0}{100.0}
               & $-$78.4
               & \matchcellbar{matchpurple}{29.4}{100.0}
               & \matchcellbar{matchgray}{9.0}{100.0}
               & $-$69.4 \\
$10{\times}10$ & \matchcellbar{matchpurple}{62.3}{100.0}
               & \matchcellbar{matchgray}{15.3}{100.0}
               & $-$75.4
               & \matchcellbar{matchpurple}{25.3}{100.0}
               & \matchcellbar{matchgray}{6.3}{100.0}
               & $-$75.1 \\
\bottomrule
\end{tabular}%
}
\caption{\textbf{Effect of removing the model's action history in Matching Pairs.} ``w/\,Act.'' keeps the action trace in context; ``w/o\,Act.'' keeps only the board images.}
\label{tab:ablation_action_feedback}
\end{minipage}
\hfill
\begin{minipage}[t]{0.48\textwidth}
\centering
\renewcommand{\arraystretch}{1.15}
\setlength\tabcolsep{4pt}
\small
\adjustbox{max width=\linewidth}{%
\begin{tabular}{l|cc|cc}
\toprule
\multirow{2}{*}{\textbf{Model Name}}
& \multicolumn{2}{c|}{\textbf{Matching Pairs}}
& \multicolumn{2}{c}{\textbf{3D Maze}} \\
& \textbf{Score\%}\,$\uparrow$ & \textbf{Resp./Score}\,$\downarrow$
& \textbf{SR\%}\,$\uparrow$ & \textbf{GS\%}\,$\uparrow$ \\
\midrule
\modelname{\qwenicon}{Qwen3.5-9B}
& \matchbar{matchpurple}{0.0}{100.0} & --
& 0.0 & \matchbar{matchpurple}{1.5}{100.0} \\
\hspace*{0.06cm}\textit{with}~\texttt{opt32k}
& \matchbar{matchpurple}{14.6}{100.0} & \matchbar{matchamber}{14.7}{130.0}
& 0.0 & \matchbar{matchpurple}{5.0}{100.0} \\
\hspace*{0.06cm}\textit{with}~\texttt{rmix32k}
& \textbf{\matchbar{matchpurple}{29.5}{100.0}} & \textbf{\matchbar{matchamber}{6.8}{130.0}}
& \textbf{10.0} & \textbf{\matchbar{matchpurple}{16.3}{100.0}} \\
\bottomrule
\end{tabular}
}
\caption{\textbf{Held-out scale evaluation of fine-tuned Qwen3.5-9B.} Evaluation sizes are strictly larger than the training data pool.}
\label{tab:sft-match-maze}
\end{minipage}
\end{table}

\noindent\textbf{More Analysis.} Additional ablations covering text-vs.-image duel observations, visual-pattern distinctiveness, the 3D Maze minimap, ask-output map externalization, and bounded-history (Markov-control) settings, together with trajectory visualizations, appear in Appendix~\ref{sec:appendix_ablation}.
\section{Training with Non-Markov Trajectories}
\label{sec:training_strategy}

Because \benchname{} is driven by two simulators rather than a fixed test set, we can roll out fresh trajectories with known optimal actions and use them as supervision. This section asks whether SFT on such rollouts teaches a smaller MLLM to act on prior observations, and is organized as: a data recipe (Sec.~\ref{sec:training_strategy}), generalization to held-out board and maze sizes (Tab.~\ref{tab:sft-match-maze}), and transfer to external memory and spatial benchmarks (Tab.~\ref{tab:qwen9b-external-transfer}).

\noindent \textbf{Setup.}
We fine-tune Qwen3.5-9B with supervised fine-tuning on action tokens, masking the loss over observation tokens.
Training trajectories are drawn from Matching Pairs boards of size $2{\times}4$ to $8{\times}8$ and 3D mazes of size $5{\times}5$ to $9{\times}9$, while all evaluation episodes use strictly larger sizes and disjoint seeds, so \textbf{no} training instance recurs at test time.

\noindent \textbf{Data construction.}
The \emph{optimal} pool is rule-based: a hand-coded oracle solves each instance in closed form, so trajectories are generated without any model, the pool scales with the number of sampled episodes, and we use 32K. The \emph{rollout} pool is harvested by running larger MLLMs (Qwen3.5-397B, Kimi-K2.5) on \benchname{} episodes and keeping only trajectories that solve the task. The correctness filter discards most rollouts, and we cap this pool at 6K. Because 6K alone is insufficient for standalone SFT in our setting, we use the rollout pool as an augmentation rather than a replacement: we compare \texttt{opt32k} (32K optimal) against \texttt{rmix32k} (26K optimal plus the 6K rollouts) at a fixed budget of 32K trajectories, so any gap isolates what the rollout component adds over an equivalently sized optimal-only baseline.

\paragraph{Held-out scale generalization.}
Tab.~\ref{tab:sft-match-maze} reports Qwen3.5-9B on board and maze sizes strictly larger than the training pool. The optimal pool alone (\texttt{opt32k}) lifts Matching Pairs from 0.0 to 14.6 and 3D Maze from 1.5 to 5.0, indicating that the supervision transfers to unseen sizes. Adding 6K rollouts (\texttt{rmix32k}) further raises Score\% to 29.5, halves response cost per matched pair, and yields the only non-zero maze SR. A plausible explanation is that the oracle pool, being mistake-free, lacks recovery states that an imperfect policy must reach; the rollout pool supplies such states.

\begingroup
\renewcommand{\arraystretch}{1.08}
\setlength\tabcolsep{5pt}
\captionsetup{hypcap=false}
\begin{center}
{%
\footnotesize
\adjustbox{max width=\columnwidth}{%
\begin{tabular}{llrrr}
\toprule
\textbf{Benchmark} & \textbf{Metric} & \textbf{Baseline} & \textbf{SFT} & $\boldsymbol{\Delta}$ \\
\midrule
\rowcolor{gray!12} \multicolumn{5}{l}{\textit{Memory- and spatial-reasoning benchmarks}} \\
\texttt{EMeMBench} & Visual DIF\_50 & 49.5 & \textbf{54.7} & \textbf{+5.2} \\
\texttt{VGRPBench} & Macro Perception Acc & 24.9 & \textbf{29.5} & \textbf{+4.6} \\
\texttt{MMSIBench\_circular} & Overall & 7.4 & \textbf{9.7} & \textbf{+2.3} \\
\texttt{ViewSpatialBench} & Overall & 41.9 & \textbf{43.4} & \textbf{+1.5} \\
\cmidrule(l){1-5}
\rowcolor{matchpurple!12} \multicolumn{2}{l}{\textit{Group mean}} & 30.9 & \textbf{34.3} & \textbf{+3.4} \\
\midrule
\rowcolor{gray!12} \multicolumn{5}{l}{\textit{General multimodal benchmarks}} \\
\texttt{OCRBench} & Final Score Norm & 85.9 & \textbf{89.2} & \textbf{+3.3} \\
\texttt{MMBench\_DEV\_EN\_V11} & Overall & 85.8 & \textbf{86.8} & +1.0 \\
\texttt{NaturalBenchDataset} & Acc & 79.7 & \textbf{80.2} & +0.5 \\
\texttt{AI2D\_TEST} & Overall & 84.6 & \textbf{85.0} & +0.4 \\
\texttt{CV-Bench-3D} & Overall & 92.7 & \textbf{93.1} & +0.4 \\
\texttt{ERQA} & Overall & 40.3 & \textbf{40.5} & +0.2 \\
\texttt{MMStar} & Overall & \textbf{73.3} & 72.7 & -0.6 \\
\texttt{RealWorldQA} & Overall & \textbf{76.6} & 75.6 & -1.0 \\
\cmidrule(l){1-5}
\rowcolor{matchpurple!12} \multicolumn{2}{l}{\textit{Group mean}} & 77.4 & \textbf{77.9} & \textbf{+0.5} \\
\bottomrule
\end{tabular}
}
}
\captionof{table}{\textbf{Results on external benchmarks.}
Scores for Qwen3.5-9B before (\textbf{Baseline}) and after (\textbf{SFT}) fine-tuning on game environment rollouts.
Benchmarks are split into a memory/spatial-reasoning group and a general group. $\Delta$ is SFT minus baseline.}
\label{tab:qwen9b-external-transfer}
\end{center}
\endgroup

\paragraph{External-benchmark transfer.}
We evaluate the same fine-tuned Qwen3.5-9B on an external suite split into a \emph{targeted} group of memory and spatial-reasoning benchmarks and a \emph{general} group covering broader multimodal abilities (Tab.~\ref{tab:qwen9b-external-transfer}). All four targeted benchmarks improve, with a group-mean gain of $+3.4$ and the largest deltas on EMeMBench ($+5.2$). On the general group the mean shifts by $+0.5$. The pattern is consistent with fine-tuning lifting the targeted capabilities without large regressions on general multimodal performance.
\section{Conclusion}
\label{sec:conclusion}

We introduced \benchname{}, a controllable benchmark for non-Markov games that isolates in-context belief-state tracking from rule understanding and perception. Across leading MLLMs, performance collapses as the latent state grows, image observations drive the bottleneck more than history length, and stripping the action trace alone reduces Matching Pairs to near-chance. Analyses localize the failure to belief-state maintenance and provide a testbed for interactive MLLMs.

\section*{Limitations}
\benchname{} focuses on two environments (Matching Pairs and 3D Maze) chosen for their controllable hidden state; broader coverage of game genres, model families, and visual styles is left to future work. In the image settings, hidden-state tracking is observed through the model's perceptual interface, and the Memory Gap metric is intended as a practical diagnostic under our oracle interface rather than a standalone causal decomposition. Our fine-tuning study is a feasibility demonstration on a single base model family.

\section*{Ethics Statement}
This work studies hidden-state tracking in controlled game environments. All visual observations are synthetically generated by our code. We do not collect human-subject data or include any personal or sensitive information, and the generated images do not depict real people or real-world copyrighted content.

\bibliographystyle{plainnat}
\bibliography{refs}


\clearpage
\appendix
\clearpage

\section{More Related Work}
\noindent\textbf{Memory Benchmarks.} Memory benchmarks target the ability to retain, update, and organize information across conversations or sessions. LoCoMo and LongMemEval evaluate long-term conversational memory through question answering over dialogue histories \citep{maharana2024locomo,wu2024longmemeval}. MemoryBench, MemBench, and MemoryAgentBench examine memory systems under continual feedback or multi-turn information accumulation \citep{ai2025memorybench,tan2025membench,hu2025memoryagentbench}. StructMemEval studies whether agents can organize memories into useful structures \citep{shutova2026structmemeval}. EMemBench is especially close in spirit: it constructs trajectory-grounded questions from text and visual game interactions to evaluate episodic memory in VLM agents \citep{li2026emembench}. We share this focus on episodic and interaction-derived memory but shift the evaluation target from answering retrospective questions to choosing prospective actions. An agent may recall a past observation in isolation yet still fail to integrate it into the latent state needed for the next move.

\noindent\textbf{Multimodal Understanding and Vision-Language Models.}
Our benchmark requires models to ground visual observations across multiple turns, a capability rooted in the rapid progress of vision-language modeling. Early few-shot multimodal learners such as Flamingo~\citep{alayrac2022flamingo} demonstrated that large-scale pre-training can bridge vision and language, and visual instruction tuning~\citep{liu2023llava} further scaled this paradigm to open-ended multimodal conversations. Recent vision-language systems have extended to richer interaction settings including free-form text-image composition, long-context streaming, and scientific reasoning~\citep{wxl_dong2024internlmxcomposer2,wxl_zhang2024omnilive,wxl_zou2026interns1pro}. On the video side, chat-centric and long-context video understanding has advanced rapidly~\citep{li2023videochat,wxl_chen2024sharegpt4video,wu2024longvideobench,fang2024mmbench, chen2025iwr}, with parallel progress on complex video object segmentation~\citep{zhang2026sec} and on probing fine-grained spatio-temporal perception~\citep{liu2026starbench}, and temporal positional encoding designs have been shown to be critical for reasoning over video frames~\citep{wxl_wei2025videorope}. Multi-turn, multi-image dialog understanding---the setting closest to our sequential observation regime---has been studied in several recent benchmarks and preference-tuning datasets~\citep{wxl_liu2024mmdu,ma2024mmlongbench,liu2025miadpo}, and fine-grained visual grounding has been pushed toward open-ended referring segmentation~\citep{zhang2026setcon} and pixel-level chart parsing~\citep{li2026visualselfrefine}.

\noindent\textbf{Reasoning, Planning, and Agent Evaluation.}
Chain-of-thought (CoT) prompting~\citep{wei2022chain,kojima2022large} has become the standard approach for eliciting multi-step reasoning in LLMs. More recent work pushes reasoning into latent or non-autoregressive spaces~\citep{nye2021show,wxl_wei2026simcot,wxl_dai2026endocot,nie2025llada,li2025beyond}, stabilizing implicit reasoning through step-level supervision or extending it to diffusion language models with variable-length denoising on structured generation tasks. A complementary line elicits \emph{visual} reasoning by ``thinking with images'' and exploiting vision--language synergy~\citep{hu2024visualsketchpad,zhang2026etchr,zhang2025thinkvisually}. On the tool-augmented side, language models have been shown to learn tool invocation autonomously~\citep{schick2023toolformer,wxl_zhang2023toolmath}, and external computation can compensate for internal reasoning limitations---a theme echoed by our external-memory interventions. For agent evaluation, the synergy between reasoning and acting~\citep{yao2023react} has motivated a growing suite of benchmarks spanning web navigation~\citep{zhou2024webarena,deng2024mind2web}, general LLM-as-agent tasks~\citep{liu2023agentbench}, computer-use environments~\citep{xie2024osworld}, and real-world long-horizon agent evaluation~\citep{wxl_ding2026wildclawbench}, with agents that self-evolve by learning from their own interaction experience~\citep{sun2025seagent}. Closely related, optimization-oriented benchmarks evaluate LLM agents over large search spaces and verifiable synthetic problems~\citep{li2025optbench,li2025npengine}, providing controllable difficulty in a different domain from our games.

\noindent\textbf{Evaluation, Alignment, and Reward Modeling of MLLMs.}
The rapid progress of multimodal models has been accompanied by a growing evaluation and alignment ecosystem~\citep{zhang2025lmmsurvey}. Open-source toolkits and broad benchmarks standardize large-scale evaluation across models and tasks~\citep{duan2024vlmevalkit,yue2024mmmu}, while dedicated benchmarks probe specific capabilities such as multimodal instruction following~\citep{zhou2023ifeval,ding2025mmifengine} and human-preference alignment~\citep{zhao2025omnialignv}. Building on reinforcement learning from human feedback and preference optimization~\citep{ouyang2022instructgpt,rafailov2023dpo} and, more recently, reinforcement learning with verifiable rewards~\citep{deepseekai2025r1}, a parallel line develops multimodal reward models and RL recipes to strengthen reasoning and alignment~\citep{yu2024rlhfv,zang2025xcomposerreward,ding2026armthinker,liu2025spark,liu2026visualerm,zhao2026trustcritic,liu2025visualrft}; verifiable rewards have further been applied to spatial understanding and dense captioning~\citep{chen2024spatialvlm,yang2026caprl,liu2026spatialssrl}. Our study is complementary: rather than measuring instruction following, preference alignment, or static capability, we isolate whether a model can maintain and act on hidden state across a long interactive history, and our fine-tuning study connects to this line by supervising on simulator rollouts with verifiable outcomes.

\section{More Analysis} \label{sec:appendix_ablation}

\noindent\textbf{Scale-sweep raw data for Fig.~\ref{fig:rq1_scale_sweep}.}
Tabs.~\ref{tab:rq1_board_size_match} and~\ref{tab:rq1_board_size_maze} report the per-size numbers underlying the scale-sweep plots in the main text. On Matching Pairs, Score\% declines from 90.6\% ($4{\times}4$) to 0.7\% ($12{\times}12$), while response cost per matched pair rises from 4.59 to 720. On 3D Maze, Game Score peaks at $7{\times}7$ (66.7\%) and drops from $9{\times}9$ onward, paralleled by declining Explore\%.

\noindent\textbf{External-memory intervention raw data for Fig.~\ref{fig:rq4_external_memory}.}
Tab.~\ref{tab:rq4_external_memory} provides the exact baseline and intervention scores behind the bar chart in the main text. On Matching Pairs ($8{\times}10$), the memory map roughly doubles Score\% for both Qwen3.5-397B (38.3\,$\to$\,78.7) and Kimi-K2.5 (43.3\,$\to$\,80.3). On 3D Maze ($9{\times}9$), the minimap yields smaller gains (Qwen: 23.8\,$\to$\,40.2; Kimi: 24.6\,$\to$\,35.6), consistent with additional bottlenecks beyond memory access.

\noindent\textbf{Oracle interface contents and MemGap reading guide.}
The Matching Pairs memory map lists, at each step, the (identity, position) pairs that have already been revealed in the current episode; positions not yet flipped are omitted. The 3D Maze minimap encodes the set of visited cells, the agent's current cell and orientation, and the wall segments observed from those cells; unvisited cells, unseen walls, and the goal location are not included. The minimap therefore reduces but does not eliminate non-Markov structure, since the agent must still plan exploration of the unobserved region. We report MemGap (Eq.~\ref{eq:memorygap}) only when the oracle score $S^*(m)$ is non-trivially above the baseline, and read its values as practical contributions of externalized state under our specific interface rather than exact decompositions of failure modes.

\renewcommand{\arraystretch}{1.15}
\begin{table}[tbp]
\setlength\tabcolsep{4.0pt}
\centering
\scriptsize
\begin{tabular}{l|ccc|ccc}
\toprule
\multirow{2}{*}{\textbf{Model Name}}
 & \multicolumn{3}{c|}{\textbf{Matching Pairs ($8{\times}10$)}}
 & \multicolumn{3}{c}{\textbf{3D Maze ($9{\times}9$)}} \\
 & \textbf{Baseline}\,$\uparrow$ & \textbf{+Memory Map}\,$\uparrow$ & \textbf{MemGap}\,$\downarrow$
 & \textbf{Baseline}\,$\uparrow$ & \textbf{+Minimap}\,$\uparrow$ & \textbf{MemGap}\,$\downarrow$ \\
\midrule
\modelname{\qwenicon}{Qwen3.5-397B}
 & \matchcellbar{matchpurple}{38.3}{100.0} & \matchcellbar{matchpurple}{78.7}{100.0} & \textbf{51.3}
 & \matchcellbar{matchpurple}{23.8}{100.0} & \matchcellbar{matchpurple}{40.2}{100.0} & \textbf{40.8} \\
\modelname{\kimiicon}{Kimi-K2.5}
 & \matchcellbar{matchpurple}{43.3}{100.0} & \matchcellbar{matchpurple}{80.3}{100.0} & 46.1
 & \matchcellbar{matchpurple}{24.6}{100.0} & \matchcellbar{matchpurple}{35.6}{100.0} & 30.9 \\
\bottomrule
\end{tabular}
\caption{\textbf{External-memory intervention on both environments.} Matching Pairs reports Score\%, 3D Maze reports GS\% (both higher is better). Matching Pairs baseline numbers are ported from the Image-Noise rows of Tab.~\ref{tab:rq2_modality}; 3D Maze baseline is the no-minimap setting. \textit{+Memory Map}: an environment-provided table of all previously revealed (position, identity) pairs. \textit{+Minimap}: an environment-provided top-down map of visited cells. \textit{MemGap}: $1 - S/S^*$ with $S$ = Baseline and $S^* =$ intervention score (Eq.~\ref{eq:memorygap}); values near $1$ localize the bottleneck to belief-state tracking.}
\label{tab:rq4_external_memory}
\end{table}

\begin{table}[tbp]
\centering
\begin{minipage}[t]{0.49\textwidth}
\centering
\renewcommand{\arraystretch}{1.08}
\setlength\tabcolsep{4pt}
\footnotesize
\adjustbox{max width=\linewidth}{%
\begin{tabular}{lccc}
\toprule
\textbf{Board} & \textbf{Pairs} & \textbf{Score\% $\uparrow$} & \textbf{Resp./Score $\downarrow$} \\
\midrule
$4{\times}4$ & 8 & \matchcellbar{matchpurple}{90.6}{100.0} & \matchcellbar{matchamber}{4.59}{40.0} \\
$4{\times}6$ & 12 & \matchcellbar{matchpurple}{56.2}{100.0} & \matchcellbar{matchamber}{8.89}{40.0} \\
$6{\times}6$ & 18 & \matchcellbar{matchpurple}{33.3}{100.0} & \matchcellbar{matchamber}{15.00}{40.0} \\
$6{\times}8$ & 24 & \matchcellbar{matchpurple}{32.3}{100.0} & \matchcellbar{matchamber}{15.48}{40.0} \\
$8{\times}8$ & 32 & \matchcellbar{matchpurple}{32.8}{100.0} & \matchcellbar{matchamber}{15.24}{40.0} \\
$8{\times}10$ & 40 & \matchcellbar{matchpurple}{23.8}{100.0} & \matchcellbar{matchamber}{21.05}{40.0} \\
$10{\times}10$ & 50 & \matchcellbar{matchpurple}{12.5}{100.0} & \matchcellbar{matchamber}{40.00}{40.0} \\
$10{\times}12$ & 60 & \matchcellbar{matchpurple}{12.1}{100.0} & \matchcellbar{matchamber}{41.38}{40.0} \\
$12{\times}12$ & 72 & \matchcellbar{matchpurple}{0.7}{100.0} & \matchcellbar{matchamber}{720.00}{40.0} \\
$12{\times}14$ & 84 & \matchcellbar{matchpurple}{3.6}{100.0} & \matchcellbar{matchamber}{140.00}{40.0} \\
\bottomrule
\end{tabular}
}
\caption{\textbf{Ablation on Matching Pairs board size.} Results are for Qwen3.5-397B in single-player action-feedback image mode with the \texttt{textures} theme and \texttt{max\_resp\_per\_pair=5}, averaged over four seeds. Resp./Score reports the response cost per matched pair.}
\label{tab:rq1_board_size_match}
\end{minipage}
\hfill
\begin{minipage}[t]{0.49\textwidth}
\centering
\renewcommand{\arraystretch}{1.08}
\setlength\tabcolsep{4.2pt}
\footnotesize
\adjustbox{max width=\linewidth}{%
\begin{tabular}{lccccc}
\toprule
\textbf{Maze} & \textbf{Cells} & \textbf{GS\% $\uparrow$} & \textbf{Eff.\% $\uparrow$} & \textbf{Explore\% $\uparrow$} \\
\midrule
$5{\times}5$   & 25  & \matchcellbar{matchpurple}{62.6}{100.0} & 46.1 & \matchcellbar{matchgray}{42}{100} \\
$7{\times}7$   & 49  & \matchcellbar{matchpurple}{66.7}{100.0} & 55.4 & \matchcellbar{matchgray}{45}{100} \\
$9{\times}9$   & 81  & \matchcellbar{matchpurple}{23.8}{100.0} & 21.9 & \matchcellbar{matchgray}{29}{100} \\
$11{\times}11$ & 121 & \matchcellbar{matchpurple}{21.8}{100.0} & 37.7 & \matchcellbar{matchgray}{20}{100} \\
$13{\times}13$ & 169 & \matchcellbar{matchpurple}{22.3}{100.0} & 51.2 & \matchcellbar{matchgray}{18}{100} \\
$15{\times}15$ & 225 & \matchcellbar{matchpurple}{19.7}{100.0} & 36.6 & \matchcellbar{matchgray}{15}{100} \\
\bottomrule
\end{tabular}
}
\caption{\textbf{Ablation on 3D Maze size.} Results are for Qwen3.5-397B in the no-minimap setting, averaged over five seeds. Cells is the maze grid size. GS\%, Eff.\%, and Explore\% are defined in Tab.~\ref{tab:main_results}.}
\label{tab:rq1_board_size_maze}
\end{minipage}

\end{table}


\noindent\textbf{Text observations dominate image observations in within-model duels, except for Seed.}
Tab.~\ref{tab:match_dual_text_vs_image_ablation} pits a text-observation player against an image-observation player from the same model family on $8{\times}8$ and $8{\times}10$ boards, with both player orders merged over three seeds. The text side wins 100\% of games for Kimi-K2.5 and Qwen3.5-397B with gaps of +22.7 to +35.7 pairs, while Seed-2.0 is the exception: the image player wins on $8{\times}8$ ($-$3.3) and the text player wins only narrowly on $8{\times}10$ (+1.8). One possible explanation is that Seed-2.0's vision encoder retains card identities more faithfully than those of Kimi and Qwen, reducing the modality gap that other models suffer from. This suggests that the text--image performance gap is not an inherent property of the task but depends on how well each model's visual pipeline preserves identity information across turns. Tab.~\ref{tab:match_dual_text_rank} aggregates the text-observation duels into the same ranking format as the main image-mode duel table for cross-modality comparison.

\renewcommand{\arraystretch}{1.08}
\begin{table}[tbp]
\setlength\tabcolsep{3.7pt}
\centering
\scriptsize
\adjustbox{max width=\textwidth}{%
\begin{tabular}{ll|ccc|ccc|rr}
\toprule
\multirow{2}{*}{\textbf{Model Pair}} & \multirow{2}{*}{\textbf{Board}}
& \multicolumn{3}{c|}{\textbf{Text Player}}
& \multicolumn{3}{c|}{\textbf{Image Player}}
& \multicolumn{2}{c}{\textbf{Outcome}} \\
& & \textbf{Model Name} & \textbf{Score} & \textbf{Win\%}
& \textbf{Model Name} & \textbf{Score} & \textbf{Win\%}
& \textbf{Gap} & \textbf{Winner} \\
\midrule
\modelname{\kimiicon}{Kimi text vs. image} & $8{\times}8$ & Kimi & 27.8 & 100.0 & Kimi & 2.8 & 0.0 & +25.0 & Text \\
\modelname{\qwenicon}{Qwen3.5 text vs. image} & $8{\times}8$ & Qwen3.5 & 27.7 & 100.0 & Qwen3.5 & 2.7 & 0.0 & +25.0 & Text \\
\modelname{\seedicon}{Seed text vs. image} & $8{\times}8$ & Seed & 7.2 & 33.3 & Seed & 10.5 & 66.7 & -3.3 & Image \\
\midrule
\modelname{\qwenicon}{Qwen3.5 text vs. image} & $8{\times}10$ & Qwen3.5 & 37.8 & 100.0 & Qwen3.5 & 2.2 & 0.0 & +35.7 & Text \\
\modelname{\kimiicon}{Kimi text vs. image} & $8{\times}10$ & Kimi & 26.7 & 100.0 & Kimi & 4.0 & 0.0 & +22.7 & Text \\
\modelname{\seedicon}{Seed text vs. image} & $8{\times}10$ & Seed & 8.0 & 33.3 & Seed & 6.2 & 50.0 & +1.8 & Text \\
\bottomrule
\end{tabular}
}
\caption{\textbf{Ablation on text versus image observations in two-player Matching Pairs.} Each row compares a text-observation player against an image-observation player under the same model family, merging both player orders over three seeds. Gap is Text score minus Image score. Missing runs are shown as --.}
\label{tab:match_dual_text_vs_image_ablation}
\end{table}

\renewcommand{\arraystretch}{1.08}
\begin{table}[tbp]
\setlength\tabcolsep{4.0pt}
\centering
\scriptsize
\adjustbox{max width=0.72\textwidth}{%
\begin{tabular}{l|cccc|c|c}
\toprule
\textbf{Model Name}
& \textbf{Elo} & \textbf{W} & \textbf{T} & \textbf{L}
& \textbf{Score\%}
& \textbf{Rank} \\
\midrule
\modelname{\qwenicon}{Qwen3.5-397B} & \textbf{1070} & 16 & 1 & 7  & \matchcellbar{matchpurple}{58.2}{100.0} & \textbf{1} \\
\modelname{\kimiicon}{Kimi-K2.5}    & 1035          & 14 & 2 & 8  & \matchcellbar{matchpurple}{56.0}{100.0} & 2 \\
\modelname{\seedicon}{Seed-2.0-Lite} & 895          & 4  & 1 & 19 & \matchcellbar{matchpurple}{33.1}{100.0} & 3 \\
\bottomrule
\end{tabular}
}
\caption{\textbf{Rank list for two-player Matching Pairs duels in text mode.} Results aggregate completed pairwise matchups across both player orders, $8{\times}8$ and $8{\times}10$ boards, and three seeds (24 games per model). Metrics follow Tab.~\ref{tab:match_dual_image_poker_rank}.}
\label{tab:match_dual_text_rank}
\end{table}


\renewcommand{\arraystretch}{1.08}
\begin{table}[tbp]
\centering
\scriptsize

\begin{minipage}[c]{0.42\textwidth}
\centering
\setlength\tabcolsep{3.5pt}
\begin{tabular}{llc}
\toprule
\textbf{Model Name} & \textbf{Theme} & \textbf{Score\%}\,$\uparrow$ \\
\midrule
\multirow{6}{*}{\modelname{\qwenicon}{Qwen3.5-397B}}
 & ASCII           & \textbf{\matchcellbar{matchpurple}{75.8}{100.0}} \\
 & Abstract        & \matchcellbar{matchpurple}{56.7}{100.0} \\
 & Similar-Colors  & \matchcellbar{matchpurple}{54.2}{100.0} \\
 & Textures        & \matchcellbar{matchpurple}{44.2}{100.0} \\
 & Noise           & \matchcellbar{matchpurple}{38.3}{100.0} \\
 & Poker           & \matchcellbar{matchpurple}{20.0}{100.0} \\
\midrule
\multirow{6}{*}{\modelname{\kimiicon}{Kimi-K2.5}}
 & ASCII           & \textbf{\matchcellbar{matchpurple}{72.5}{100.0}} \\
 & Similar-Colors  & \matchcellbar{matchpurple}{70.8}{100.0} \\
 & Textures        & \matchcellbar{matchpurple}{50.8}{100.0} \\
 & Noise           & \matchcellbar{matchpurple}{43.3}{100.0} \\
 & Abstract        & \matchcellbar{matchpurple}{41.4}{100.0} \\
 & Poker           & \matchcellbar{matchpurple}{30.1}{100.0} \\
\bottomrule
\end{tabular}
\end{minipage}%
\hfill
\begin{minipage}[c]{0.56\textwidth}
\centering
\setlength\tabcolsep{3.0pt}
\adjustbox{max width=\linewidth}{%
\begin{tabular}{lc cccc cccc}
\toprule
\textbf{Model Name} & \textbf{Size}
 & \multicolumn{4}{c}{\textbf{Baseline}}
 & \multicolumn{4}{c}{\textbf{Color Tag}} \\
\cmidrule(lr){3-6} \cmidrule(lr){7-10}
& & SR & Eff.\% & Explore\% & GS\%
& SR & Eff.\% & Explore\% & GS\% \\
\midrule
\multirow{3}{*}{\modelname{\kimiicon}{Kimi-K2.5}}
& $7{\times}7$
& 1/5 & 66.7 & 44.0 & \matchcellbar{matchpurple}{34.3}{100.0}
& \textbf{4/5} & 39.9 & 47.0 & \textbf{\matchcellbar{matchpurple}{60.7}{100.0}} \\
& $9{\times}9$
& \textbf{2/5} & 62.5 & 24.0 & \textbf{\matchcellbar{matchpurple}{39.7}{100.0}}
& 1/5 & 100.0 & 19.0 & \matchcellbar{matchpurple}{27.6}{100.0} \\
& $11{\times}11$
& 2/5 & 68.1 & 25.0 & \textbf{\matchcellbar{matchpurple}{41.1}{100.0}}
& 2/5 & 33.5 & 24.0 & \matchcellbar{matchpurple}{33.9}{100.0} \\
\midrule
\multirow{3}{*}{\modelname{\qwenicon}{Qwen3.5-397B}}
& $7{\times}7$
& 4/5 & 55.4 & 47.0 & \textbf{\matchcellbar{matchpurple}{66.9}{100.0}}
& 4/5 & 30.0 & 51.0 & \matchcellbar{matchpurple}{57.1}{100.0} \\
& $9{\times}9$
& 1/5 & 21.9 & 29.0 & \matchcellbar{matchpurple}{23.8}{100.0}
& \textbf{2/5} & 84.3 & 24.0 & \textbf{\matchcellbar{matchpurple}{44.1}{100.0}} \\
& $11{\times}11$
& 1/5 & 37.7 & 20.0 & \matchcellbar{matchpurple}{21.8}{100.0}
& \textbf{2/5} & 32.2 & 27.0 & \textbf{\matchcellbar{matchpurple}{34.5}{100.0}} \\
\bottomrule
\end{tabular}
}
\end{minipage}

\caption{\textbf{Visual-pattern ablation.} \textit{Left}: Matching Pairs theme sweep (with action feedback), varying card-identifier distinctiveness while keeping rules fixed. \textit{Right}: 3D Maze baseline (uniform walls) vs.\ Color Tag (wall segments painted with distinct colors as visual landmarks). Identifier distinctiveness strongly affects Matching Pairs, while wall color tags do not consistently help 3D Maze.}
\label{tab:rq3_visual_pattern}
\end{table}

\noindent\textbf{Visual identifier distinctiveness strongly affects Matching Pairs, but wall color tags do not consistently help 3D Maze.} We test whether failures arise only from long-horizon memory or also from unstable visual identity binding. Matching Pairs ($8{\times}10$) varies card-pattern distinctiveness across six themes; 3D Maze compares uniform walls against color-tagged walls. Tab.~\ref{tab:rq3_visual_pattern} reports both. On Matching Pairs, Qwen3.5-397B drops from 75.8\% (ASCII) to 20.0\% (Poker), and Kimi-K2.5 from 72.5\% to 30.1\%, showing that less distinctive identifiers compound the memory load. On 3D Maze the color-tag intervention is mixed: Kimi-K2.5 improves at $7{\times}7$ (1/5\,$\to$\,4/5) but degrades at $9{\times}9$ (2/5\,$\to$\,1/5), and Qwen3.5-397B gains modestly at $9{\times}9$ and $11{\times}11$. Hidden-state tracking is therefore coupled with perceptual binding for card identification, while wall-level visual landmarks alone fail to stabilize spatial belief updates. The asymmetry between the two environments is informative: in Matching Pairs, each card identity must be discriminated from dozens of visually similar alternatives, so pattern distinctiveness directly affects recognition accuracy. In 3D Maze, wall colors provide only coarse spatial cues that do not resolve the core difficulty of maintaining a global position estimate from local egocentric views. This distinction suggests that visual interventions are most effective when they target the specific perceptual bottleneck of the task.






\noindent\textbf{The minimap helps mostly at intermediate and large scales, and its effect is model-dependent.}
Tab.~\ref{tab:ablation_minimap} reports the minimap condition across $5{\times}5$ to $15{\times}15$ mazes with exploration and collision statistics. Seed-2.0-Pro benefits the most, reaching 3/5 success at both $9{\times}9$ and $11{\times}11$ and recovering 2/5 at $15{\times}15$. At $13{\times}13$, Qwen3.5-397B and Seed-2.0-Pro each complete one minimap run, while Seed-2.0-Lite and Kimi-K2.5 fail on all seeds. The Walls column reveals a second pattern: Kimi-K2.5 keeps collisions low even when it fails, showing that safer local control does not guarantee global completion. This dissociation between local safety and global success highlights two distinct failure modes in 3D Maze: models can fail either because they misread the immediate 3D scene (leading to wall collisions) or because they lose track of the global spatial layout (leading to loops and revisits). Kimi-K2.5 predominantly exhibits the latter failure, maintaining accurate local perception while lacking a coherent spatial map.

\renewcommand{\arraystretch}{1.1}
\begin{table}[tbp]
\setlength\tabcolsep{3.0pt}
\centering
\scriptsize
\adjustbox{max width=\textwidth}{%
\begin{tabular}{lccc ccc ccc ccc ccc ccc}
\toprule
\textbf{Model Name}
 & \multicolumn{3}{c}{\textbf{$5{\times}5$}}
 & \multicolumn{3}{c}{\textbf{$7{\times}7$}}
 & \multicolumn{3}{c}{\textbf{$9{\times}9$}}
 & \multicolumn{3}{c}{\textbf{$11{\times}11$}}
 & \multicolumn{3}{c}{\textbf{$13{\times}13$}}
 & \multicolumn{3}{c}{\textbf{$15{\times}15$}} \\
\cmidrule(lr){2-4} \cmidrule(lr){5-7} \cmidrule(lr){8-10} \cmidrule(lr){11-13} \cmidrule(lr){14-16} \cmidrule(lr){17-19}
& Eff.\% & Explore\% & GS\%
& Eff.\% & Explore\% & GS\%
& Eff.\% & Explore\% & GS\%
& Eff.\% & Explore\% & GS\%
& Eff.\% & Explore\% & GS\%
& Eff.\% & Explore\% & GS\% \\
\midrule
\modelname{\seedicon}{Seed-2.0-Lite} & 36.9 & 46.0 & 59.4
& 31.6 & 40.0 & 29.2
& 39.0 & 26.0 & 35.6
& 0.0 & 16.0 & 8.0
& 0.0 & 16.0 & 8.0
& 0.0 & 10.0 & 5.0 \\
\modelname{\kimiicon}{Kimi-K2.5} & 61.8 & 43.0 & 69.0
& 41.5 & 36.0 & 39.1
& 41.2 & 32.0 & 37.8
& 36.5 & 26.0 & 35.1
& 0.0 & 19.0 & 9.5
& 0.0 & 14.0 & 7.0 \\
\modelname{\qwenicon}{Qwen3.5-397B} & 58.8 & 44.0 & 79.4
& 33.5 & 46.0 & 58.0
& 54.7 & 31.0 & 40.2
& 34.4 & 28.0 & 45.9
& 33.6 & 23.0 & 22.6
& 0.0 & 13.0 & 6.5 \\
\modelname{\seedicon}{Seed-2.0-Pro} & 56.6 & 48.0 & 78.3
& 69.9 & 43.0 & 59.6
& 70.6 & 34.0 & 58.0
& 63.0 & 30.0 & 54.9
& 75.9 & 20.0 & 25.6
& 51.1 & 22.0 & 36.8 \\
\bottomrule
\end{tabular}
}
\caption{\textbf{Maze-size sweep with minimap.} Eff.\%, Explore\%, and GS\% are defined in Tab.~\ref{tab:main_results}; GS\% (aggregate) is rightmost in each size block.}
\label{tab:ablation_minimap}
\end{table}

\noindent\textbf{Ask-output prompting helps Seed-2.0-Lite substantially but does not improve overall completion for Kimi or Qwen.}\label{sec:appendix_askmap}
Tab.~\ref{tab:ablation_askmap} reports results when the agent is prompted to explicitly output its internal spatial map at each step. Seed-2.0-Lite rises from 1/5 to 4/5, while Kimi-K2.5 and Qwen3.5-397B show no gain in completion. Their successful ask-output runs are near-optimal (Eff = 1.000) when they work, but occur no more frequently than in the standard setting. Seed-2.0-Pro maintains 3/5 success in both conditions with slightly higher efficiency under ask-output. The Traj-Match column reveals that map-output quality varies widely: Seed-2.0-Pro reaches 87.2\% and Qwen3.5-397B achieves 82.7\%, while Seed-2.0-Lite and Kimi-K2.5 score lower (66.3\% and 68.2\%). Explicit map externalization thus benefits models that can maintain accurate spatial representations. The gap between Traj-Match accuracy and task completion suggests that merely producing a map is not enough: the model must also use that map to plan. A model with high Traj-Match but low success rate accurately records where it has been yet fails to translate that record into effective route planning, consistent with a decision-selection contribution in addition to memory access.
\renewcommand{\arraystretch}{1.1}
\begin{table}[tbp]
\setlength\tabcolsep{4.0pt}
\centering
\scriptsize
\adjustbox{max width=\textwidth}{%
\begin{tabular}{lccc cccc}
\toprule
\textbf{Model Name}
 & \multicolumn{3}{c}{\textbf{Standard}}
 & \multicolumn{4}{c}{\textbf{Ask-Output}} \\
\cmidrule(lr){2-4} \cmidrule(lr){5-8}
& Eff.\% & Explore\% & GS\%
& Eff.\% & Explore\% & Traj-Match\% & GS\% \\
\midrule
\modelname{\kimiicon}{Kimi-K2.5}    & 62.5 & 23.0 & 39.4 & 100.0 & 28.0 & 68.2 & 31.2 \\
\modelname{\qwenicon}{Qwen3.5-397B} & 21.9 & 29.0 & 23.8 & 100.0 & 19.0 & 82.7 & 27.6 \\
\modelname{\seedicon}{Seed-2.0-Lite} & 26.2 & 30.0 & 24.6 & 49.9 & 35.0 & 66.3 & 63.5 \\
\modelname{\seedicon}{Seed-2.0-Pro}  & 60.2 & 32.0 & 54.5 & 67.2 & 31.0 & 87.2 & 56.4 \\
\bottomrule
\end{tabular}
}
\caption{\textbf{Ask-output ablation on $9{\times}9$ mazes.} The ask-output setting explicitly asks the model to emit its internal wall map at every step; Traj-Match measures trajectory-map agreement.}
\label{tab:ablation_askmap}
\end{table}

\noindent\textbf{Limiting context does not affect all models equally, and more history does not uniformly help.}\label{sec:appendix_markov}
Tab.~\ref{tab:ablation_markov} compares the standard full-history setting with history windows of 3, 5, and 10 turns on the extended $9{\times}9$ experiment. Seed-2.0-Pro achieves its best success (3/5) under full history and drops to 2/5 across all three windows, consistent with exploiting long-range dependencies. Qwen3.5-397B shows the opposite trend: only 1/5 with full history but 2/5 under all windowed conditions. Very long dialogue history may impair its decisions through attention dilution or accumulated misinformation. Kimi-K2.5 shows a milder version of this pattern (2/5 under full history, dropping to 1/5 at windows of 5 and 10), and Seed-2.0-Lite remains unstable across all conditions (1--2/5). The history--performance relationship is model-dependent: some models benefit from richer context while others are hindered by it.

\renewcommand{\arraystretch}{1.1}
\begin{table}[tbp]
\setlength\tabcolsep{3.2pt}
\centering
\scriptsize
\adjustbox{max width=\textwidth}{%
\begin{tabular}{lccc ccc ccc ccc}
\toprule
\textbf{Model Name}
 & \multicolumn{3}{c}{\textbf{Full History}}
 & \multicolumn{3}{c}{\textbf{Window = 3}}
 & \multicolumn{3}{c}{\textbf{Window = 5}}
 & \multicolumn{3}{c}{\textbf{Window = 10}} \\
\cmidrule(lr){2-4} \cmidrule(lr){5-7} \cmidrule(lr){8-10} \cmidrule(lr){11-13}
& Eff.\% & Explore\% & GS\%
& Eff.\% & Explore\% & GS\%
& Eff.\% & Explore\% & GS\%
& Eff.\% & Explore\% & GS\% \\
\midrule
\modelname{\kimiicon}{Kimi-K2.5}    & 62.5 & 23.0 & 39.4 & 66.1 & 26.0 & 41.0 & 93.3 & 28.0 & 30.5 & 100.0 & 25.0 & 30.0 \\
\modelname{\qwenicon}{Qwen3.5-397B} & 21.9 & 29.0 & 23.8 & 56.3 & 30.0 & 40.3 & 26.6 & 31.0 & 34.6 & 58.9 & 33.0 & 41.7 \\
\modelname{\seedicon}{Seed-2.0-Lite} & 26.2 & 30.0 & 24.6 & 82.4 & 34.0 & 31.8 & 25.8 & 33.0 & 35.1 & 73.7 & 23.0 & 26.6 \\
\modelname{\seedicon}{Seed-2.0-Pro}  & 60.2 & 32.0 & 54.5 & 42.6 & 21.0 & 34.8 & 64.5 & 30.0 & 41.9 & 64.6 & 31.0 & 42.2 \\
\bottomrule
\end{tabular}
}
\caption{\textbf{History-window ablation on $9{\times}9$ mazes without minimap.} Full history is compared against windows retaining only the most recent 3, 5, or 10 turns.}
\label{tab:ablation_markov}
\end{table}



\section{Visualization Cases}

The following trajectory visualizations illustrate three common failure modes observed across our experiments: (i) \emph{spatial drift}, where the model's stated position gradually diverges from its actual location; (ii) \emph{local oscillation}, where the model revisits the same small region without making global progress; and (iii) \emph{reasoning--action mismatch}, where the model's verbal reasoning is spatially coherent but its chosen action contradicts its own stated plan. Each figure contrasts a successful and a failed run on the same maze to highlight where the reasoning diverges.

\begin{figure}[tbp]
\centering
\includegraphics[width=\textwidth]{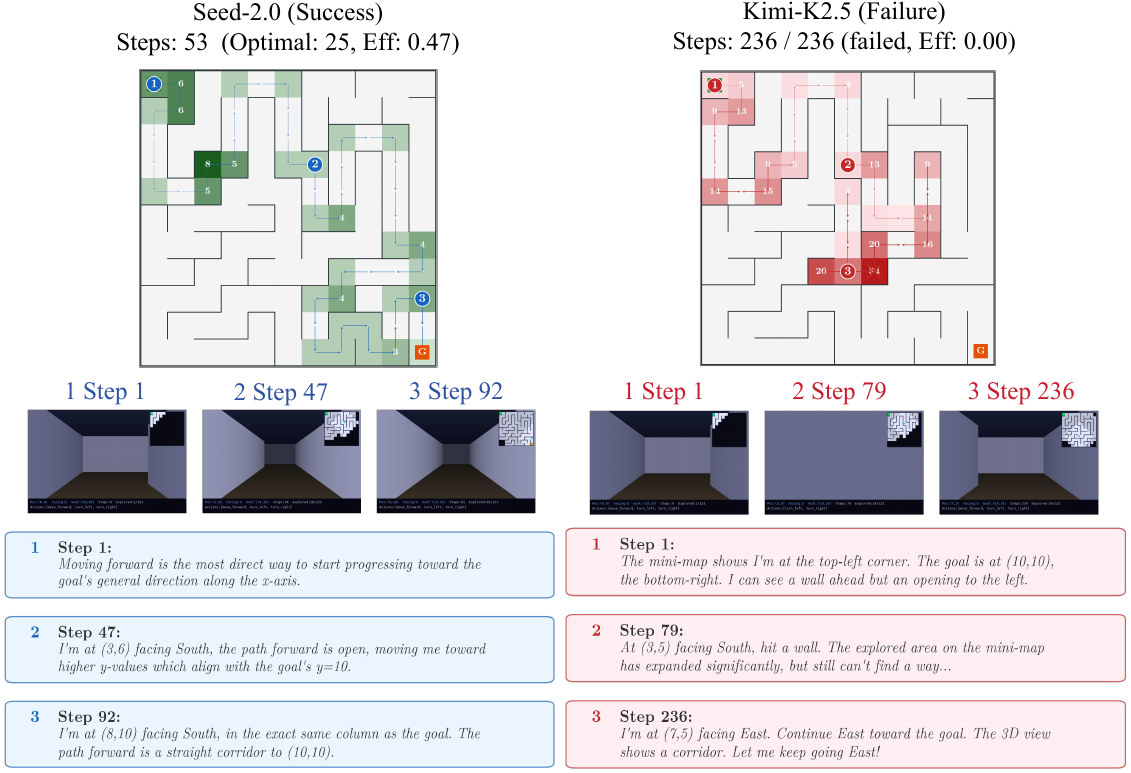}
\caption{Success rate with minimap across maze sizes, corresponding to Tab.~\ref{tab:ablation_minimap}.}
\label{fig:ab_minimap}
\end{figure}

\noindent\textbf{Successful runs maintain spatially grounded reasoning; failed runs drift into confusion despite the minimap.}\label{sec:appendix_viz_minimap}
Fig.~\ref{fig:ab_minimap} contrasts two 3D Maze trajectories under the minimap condition. Seed-2.0 reaches the goal in 53 steps (optimal 25, Eff 0.47); its reasoning anchors each move to explicit coordinates and headings (``at $(3,6)$ facing South'', ``straight corridor to $(10,10)$''), so the trajectory (green) advances steadily toward the goal. Kimi-K2.5 exhausts the 236-step budget on the same map without reaching the goal. Its reasoning still cites the minimap but loses spatial closure (``hit a wall, the explored area has expanded significantly, but still can't find a way'') and oscillates around the lower-right region (red trajectory). Providing an external minimap does not by itself fix hidden-state tracking: the model must still translate the minimap into a consistent position belief. Failures appear as repeated wall collisions and incoherent direction choices rather than as missing information. This case demonstrates that the minimap's value depends on the model's ability to ground its position within the provided map. When this grounding fails, the minimap becomes decorative: the model cites it in its reasoning but cannot use it to correct accumulated spatial errors.

\begin{figure}[tbp]
\centering
\includegraphics[width=\textwidth]{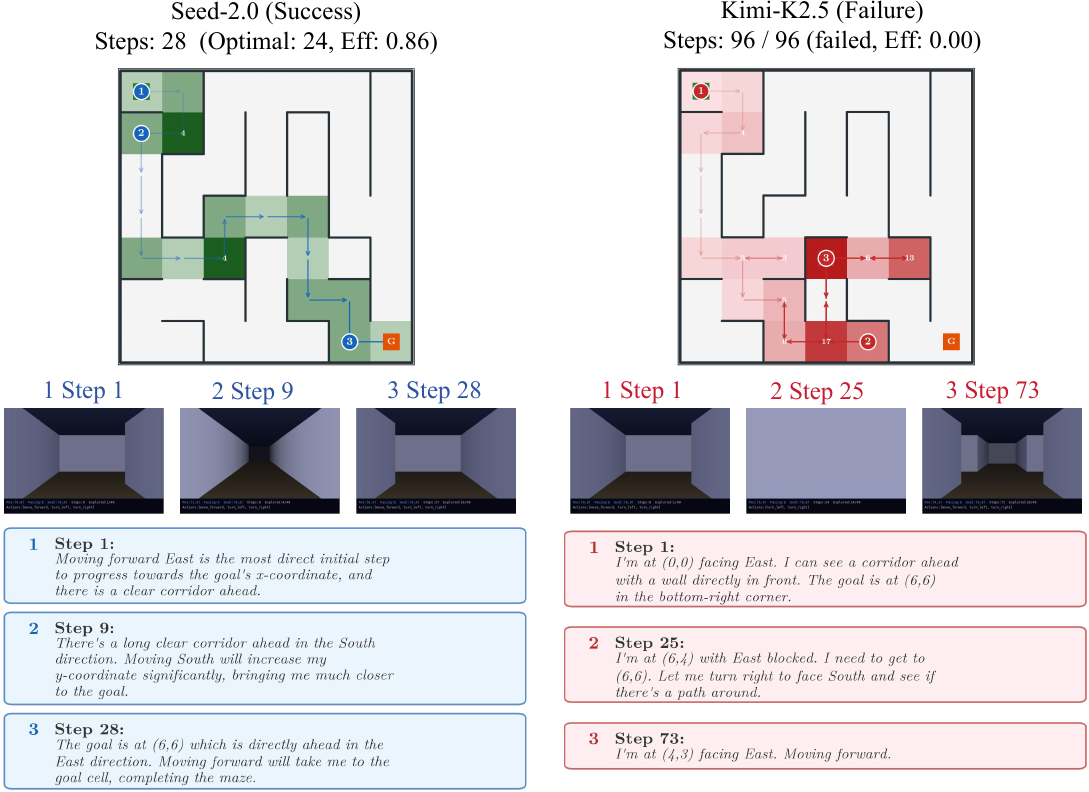}
\caption{\textbf{Baseline 3D Maze trajectories (no minimap, uniform walls).} Seed-2.0 (left, blue) reaches the goal in 28 steps (optimal 24, Eff 0.86) with spatially grounded reasoning. Kimi-K2.5 (right, red) exhausts the 96-step budget, revisiting cells and oscillating despite recognizing dead ends.}
\label{fig:ab_vanilla}
\end{figure}

\noindent\textbf{Without any external aid, strong spatial grounding separates success from failure.}
Fig.~\ref{fig:ab_vanilla} shows a $7{\times}7$ baseline run (no minimap, uniform walls). Seed-2.0 maintains coordinate-level awareness throughout (``moving South increases my y-coordinate, bringing me closer to the goal'') and reaches the goal near-optimally. Kimi-K2.5 correctly identifies dead ends (``I need a completely different approach, I've been going in circles'') but lacks the spatial map to act on this realization, resulting in repeated revisits. The contrast reveals that recognizing failure is not the bottleneck: Kimi-K2.5 diagnoses the problem in natural language but cannot translate that diagnosis into a corrective spatial plan. Without a persistent internal map, each ``different approach'' amounts to a random direction change rather than a systematic exploration strategy.

\begin{figure}[tbp]
\centering
\includegraphics[width=\textwidth]{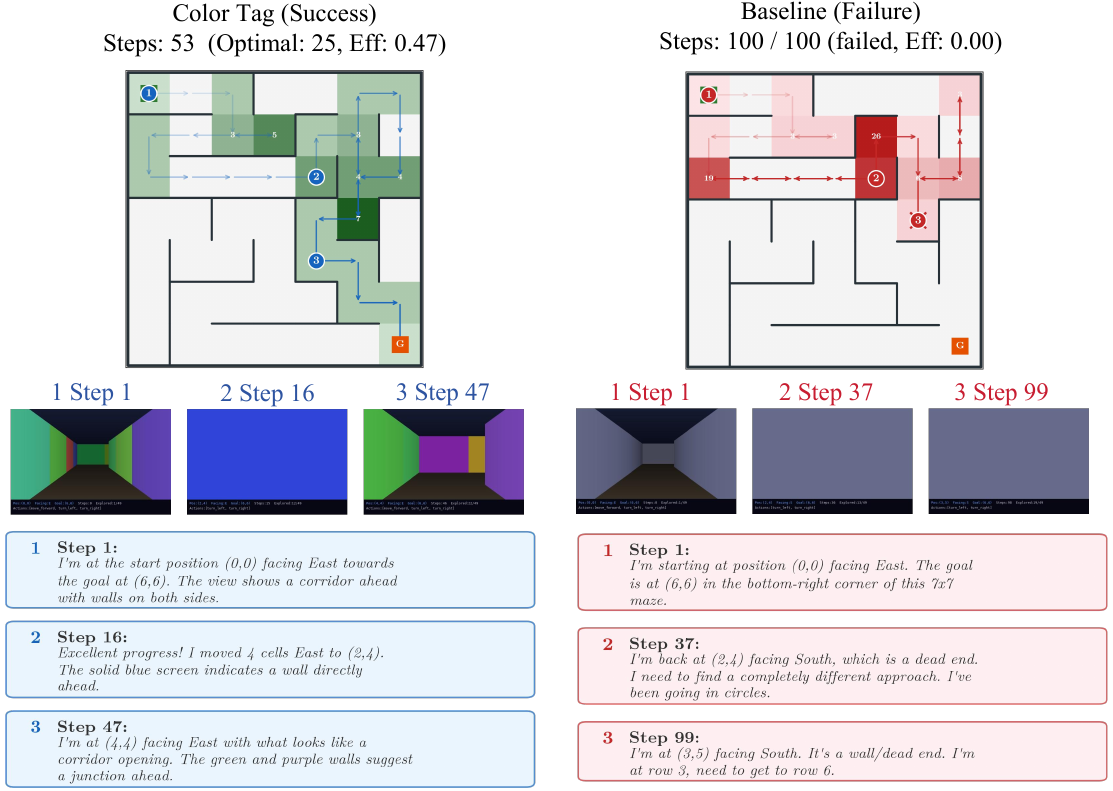}
\caption{\textbf{Color Tag vs.\ Baseline on 3D Maze.} Left (blue): color-tagged walls help Seed-2.0 navigate a $7{\times}7$ maze in 53 steps (Eff 0.47). Right (red): the same model fails under uniform walls, exhausting the 100-step budget with repeated oscillations.}
\label{fig:ab_pattern}
\end{figure}

\noindent\textbf{Color-tagged walls can help spatial disambiguation but do not eliminate failures.}
Fig.~\ref{fig:ab_pattern} compares trajectories on the same $7{\times}7$ maze under two visual conditions. With color tags, Seed-2.0 anchors its reasoning to wall colors and junction landmarks, reaching the goal in 53 steps. Under uniform walls, the same model loses spatial orientation after step 26 and oscillates in the upper-right region until timeout. The color tags provide perceptual anchors that stabilize position tracking, but as shown in Tab.~\ref{tab:rq3_visual_pattern}, this advantage does not generalize across all models or maze sizes. Across the three visualization cases, a consistent pattern emerges: successful models maintain an explicit, coordinate-level spatial representation that they update after each action, while failed models rely on qualitative spatial language (``keep going'', ``try a different direction'') that does not accumulate into a coherent map. The key differentiator is not whether the model can perceive the current view, but whether it can integrate that view with its history to form a stable belief about global position.

\section{Matching Pairs Case Studies}

\begin{figure}[tbp]
\centering
\includegraphics[width=\textwidth]{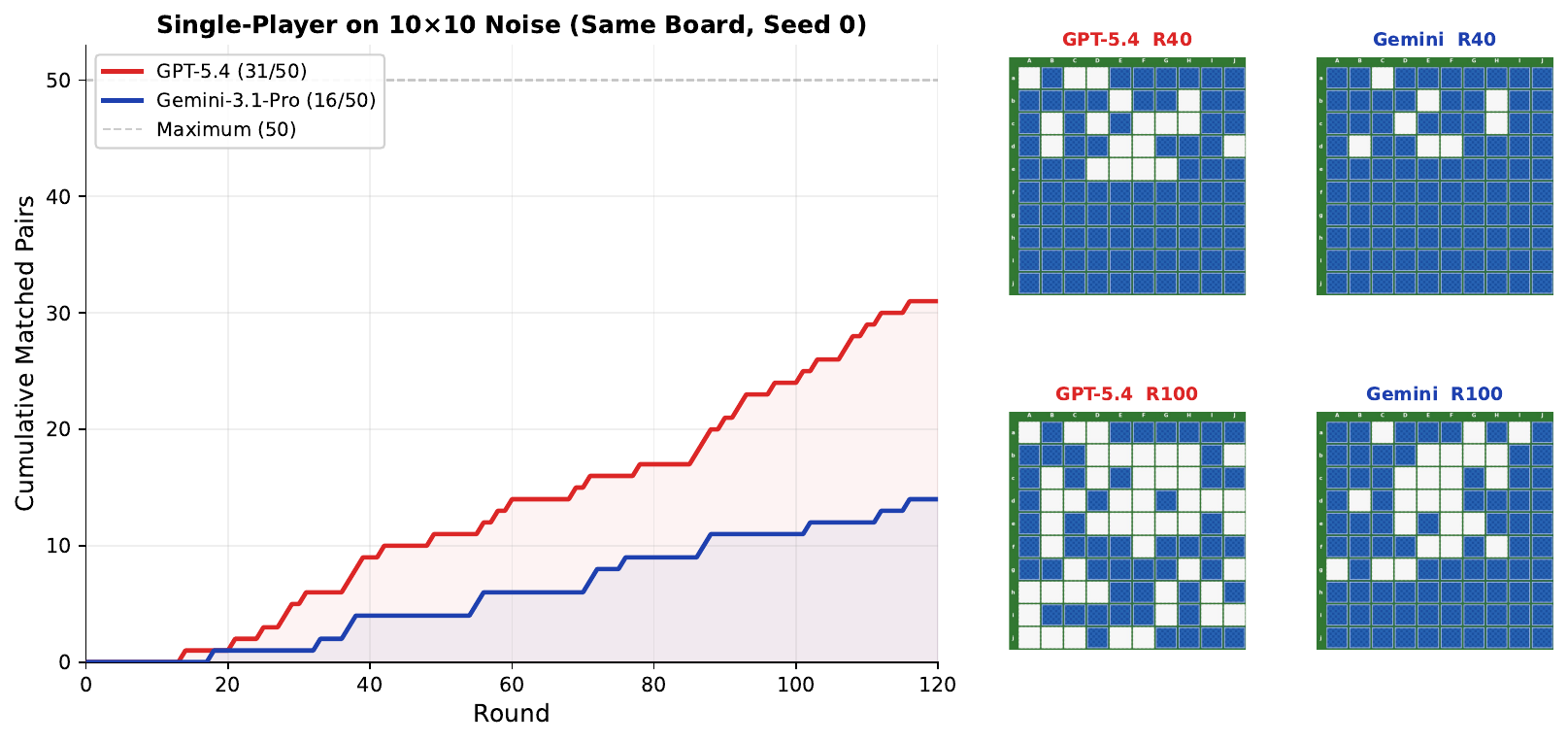}
\caption{\textbf{Single-player Matching Pairs trajectory.} GPT-5.4 and Gemini-3.1-Pro are evaluated on the same $10{\times}10$ noise board with seed 0. The curve reports cumulative matched pairs over 120 rounds, and the snapshots show board states at selected rounds.}
\label{fig:case_single_player_matching}
\end{figure}

\paragraph{Single-player trajectory.}
Fig.~\ref{fig:case_single_player_matching} compares GPT-5.4 and Gemini-3.1-Pro on the same $10{\times}10$ noise board. GPT-5.4 finishes with 31 of 50 pairs, while Gemini-3.1-Pro reaches 16 of 50. Beyond the final score, the trajectory reveals a clear difference in how observations are converted into matches. Gemini-3.1-Pro shows several plateau phases, where continued flips do not lead to a score. This indicates a weaker ability to maintain and reuse card-location bindings after they disappear from the current observation. GPT-5.4 instead shows a sharp improvement between rounds 80 and 100. This late jump suggests that earlier observations have been integrated into a more useful belief state, which then supports a concentrated phase of successful matching. The snapshots are consistent with this pattern, as GPT-5.4 clears a larger region of the board by round 100 while Gemini-3.1-Pro leaves more cells unresolved. This case highlights that the benchmark tests not only visual observation but also whether the model can transform past observations into a stable hidden-state estimate for later action.

\begin{figure}[tbp]
\centering
\includegraphics[width=\textwidth]{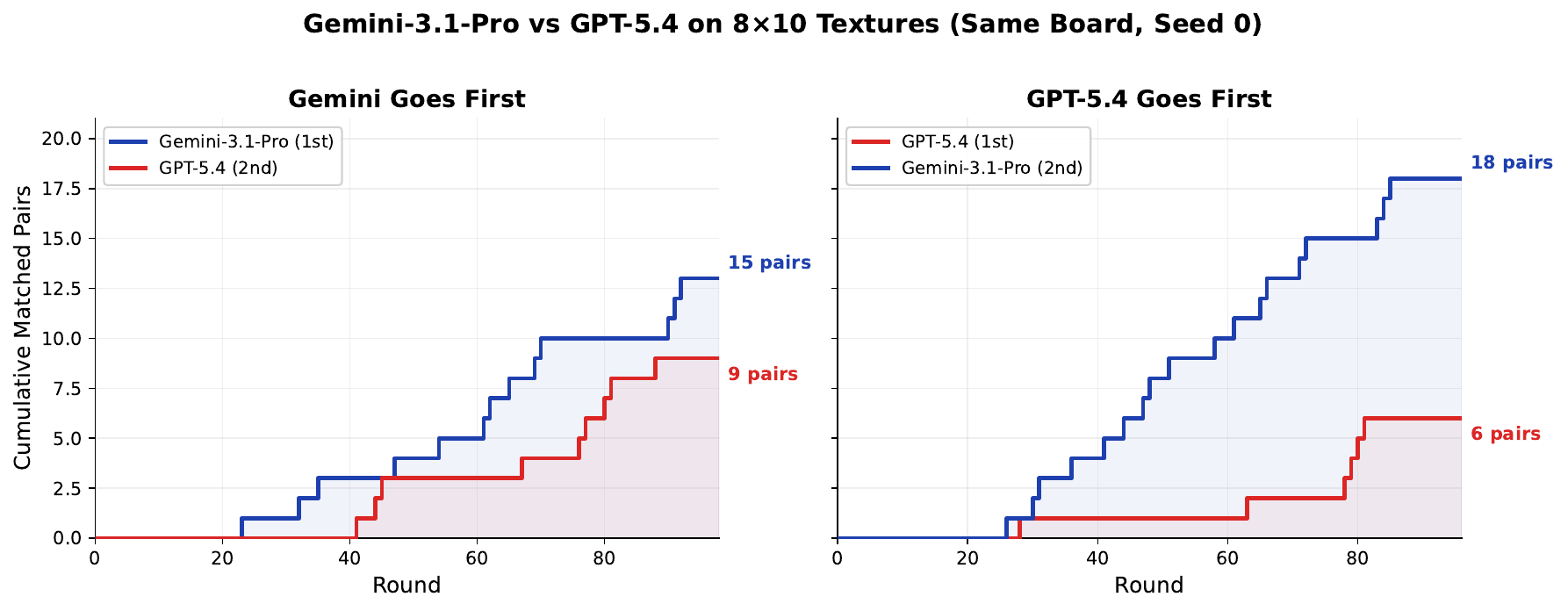}
\caption{\textbf{Duel Matching Pairs trajectory.} Gemini-3.1-Pro and GPT-5.4 are compared on the same $8{\times}10$ texture board with seed 0 under both player orders. The two panels test whether the outcome is mainly caused by first-move advantage.}
\label{fig:case_dual_matching}
\end{figure}

\paragraph{Duel trajectory.}
Fig.~\ref{fig:case_dual_matching} compares the two models on the same $8{\times}10$ texture board under both player orders. When Gemini-3.1-Pro moves first, it finishes with 15 pairs, while GPT-5.4 reaches 9. When GPT-5.4 moves first, Gemini-3.1-Pro still finishes with 18 pairs, while GPT-5.4 reaches 6. This suggests that the duel result is not mainly driven by first-move advantage. Gemini-3.1-Pro gains more matches in the late game and appears to use earlier observations more effectively. The duel setting therefore provides a stricter test of internal belief state reconstruction, because each model must remember both its own observations and the cards revealed by the opponent.

\section{Potential Risks}
The main risk of this work is possible over-interpretation of benchmark results as a complete measure of model intelligence or real-world reliability. To reduce this risk, we report the task settings, evaluation metrics, and limitations of our benchmark, and we frame the results as evidence about hidden-state tracking under controlled conditions. The benchmark is intended for research evaluation and diagnostic analysis, rather than for direct deployment decisions in safety-critical applications.

\section{SFT Training Details}\label{sec:appendix_sft}

All Qwen3.5-9B fine-tuning runs in §\ref{sec:training_strategy} use the same recipe; only the training data composition changes across the \texttt{opt32k} and \texttt{rmix32k} configurations.

\noindent\textbf{Finetuning scope.} We perform full fine-tuning of the language model with the vision tower and multimodal projector frozen, using the \texttt{qwen3\_5\_nothink} chat template.

\noindent\textbf{Optimization.} AdamW with learning rate $1\mathrm{e}{-5}$, cosine schedule, warmup ratio $0.1$, 1 epoch, and an effective batch size of 128 (per-device batch size 1 with gradient accumulation 16 across 8 GPUs).

\noindent\textbf{Sequence and visual budget.} Cutoff length 28{,}160 tokens for \texttt{opt32k} and 29{,}000 tokens for \texttt{rmix32k}; per-image maximum 65{,}536 pixels with patch cropping enabled.

\noindent\textbf{Data composition.} \texttt{opt32k} mixes 16k Matching Pairs and 16k 3D Maze optimal-policy rollouts ($\sim$32k records total); \texttt{rmix32k} replaces each 16k optimal block with a 16k rollout-plus-optimal mixture from the same environment. The two datasets are concatenated with the \texttt{concat} mix strategy.

\noindent\textbf{Reproducibility.} Random seed and other unspecified hyperparameters follow LLaMA-Factory defaults. Each configuration is trained with a single seed; downstream evaluation aggregates over 5 environment seeds per configuration (see §\ref{sec:metrics}).


\end{document}